\documentclass[preprint]{elsarticle}
\usepackage[utf8]{inputenc}
\usepackage{float} 
\usepackage{amsmath}
\usepackage{multirow}
\usepackage[table,xcdraw]{xcolor}
\usepackage{graphicx}
\usepackage{caption}
\PassOptionsToPackage{hyphens}{url}\usepackage{hyperref}
\usepackage{hyperref}

\journal{Artificial Intelligence in Medicine}

\begin{document}

\begin{frontmatter}

\title{A New Approach for Interpretability and Reliability in Clinical Risk Prediction: Acute Coronary Syndrome Scenario\tnoteref{t1}}
\tnotetext[t1]{© 2021. This manuscript version is made available under the CC-BY-NC-ND 4.0 license http://creativecommons.org/licenses/by-nc-nd/4.0/}

\renewcommand*{\today}{}

\let\today\relax
\makeatletter
\def\ps@pprintTitle{%
    \let\@oddhead\@empty
    \let\@evenhead\@empty
    \def\@oddfoot{\footnotesize\itshape
         {Accepted for publication in Artificial Intelligence in Medicine \par
         (10.1016/j.artmed.2021.102113)} \hfill\today}%
    \let\@evenfoot\@empty
    }
\makeatother

\author[1]{Francisco Valente\corref{cor1}}
\ead{paulo.francisco.valente@gmail.com}
\author[1]{Jorge Henriques}
\ead{jh@dei.uc.pt}
\author[2,1]{Simão Paredes}
\ead{sparedes@isec.pt}
\author[2,1]{Teresa Rocha}
\ead{teresa@isec.pt}
\author[1]{Paulo de Carvalho}
\ead{carvalho@dei.uc.pt}
\author[3]{João Morais}
\ead{joaomorais@chleiria.min-saude.pt}

\cortext[cor1]{Corresponding author}

\address[1]{Center for Informatics and Systems of University of Coimbra, University of Coimbra, Pólo II, 3030-290 Coimbra, Portugal}
\address[2]
{Polytechnic of Coimbra, Department of Systems and Computer Engineering, Rua Pedro Nunes - Quinta da Nora, 3030-199 Coimbra, Portugal}
\address[3]{Cardiology Department, Leiria Hospital Centre, Leiria, Portugal}

\begin{abstract}
\textit{Introduction:} The risk prediction of the occurrence of a clinical event is often based on conventional statistical procedures, through the implementation of risk score models. Recently, approaches based on more complex machine learning (ML) methods have been developed. Despite the latter usually have a better predictive performance, they obtain little approval from the physicians, as they lack interpretability and, therefore, clinical confidence. One clinical issue where both types of models have received great attention is the mortality risk prediction after acute coronary syndromes (ACS).
\par
\textit{Objective:} We intend to create a new risk assessment methodology that combines the best characteristics of both risk score and ML models. More specifically, we aim to develop a method that, besides having a good performance, offers a personalized model and outcome for each patient, presents high interpretability, and incorporates an estimation of the prediction reliability which is not usually available. By combining these features in the same approach we expect that it can boost the confidence of physicians to use such a tool in their daily activity.
\par
\textit{Methods:} In order to achieve the mentioned goals, a three-step methodology was developed: several rules were created by dichotomizing risk factors; such rules were trained with a machine learning classifier to predict the acceptance degree of each rule (the probability that the rule is correct) for each patient; that information was combined and used to compute the risk of mortality and the reliability of such prediction. The methodology was applied to a dataset of 1111 patients admitted with any type of ACS (myocardial infarction and unstable angina) in two Portuguese hospitals, to assess the 30-days all-cause mortality risk, being validated through a Monte-Carlo cross-validation technique. The performance was compared with state-of-the-art approaches: logistic regression (LR), artificial neural network (ANN), and clinical risk score model (namely the Global Registry of Acute Coronary Events - GRACE).
\par
\textit{Results:} For the scenario being analyzed, the performance of the proposed approach and the comparison models was assessed through discrimination and calibration. The ability to rank the patients was evaluated through the area under the ROC curve (AUC), and the ability to stratify the patients into low or high-risk groups was determined using the geometric mean (GM) of specificity and sensitivity, the negative predictive value (NPV) and the positive predictive value (PPV). The validation calibration curves were also inspected. The proposed approach (AUC=81\%, GM=74\%, PPV=17\%, NPV=99\%) achieved testing results identical to the standard LR model (AUC=83\%, GM=73\%, PPV=16\%, NPV=99\%), but offers superior interpretability and personalization; it also significantly outperforms the GRACE risk model (AUC=79\%, GM=47\%, PPV=13\%, NPV=98\%) and the standard ANN model (AUC=78\%, GM=70\%, PPV=13\%, NPV=98\%). The calibration curve also suggests a very good generalization ability of the obtained model as it approaches the ideal curve (slope=0.96). Finally, the reliability estimation of individual predictions presented a great correlation with the misclassifications rate.
\par
\textit{Conclusion:} We developed and described a new tool that showed great potential to guide the clinical staff in the risk assessment and decision-making process, and to obtain their wide acceptance due to its interpretability and reliability estimation properties.  The methodology presented a good performance when applied to ACS events, but those properties may have a beneficial application in other clinical scenarios as well.

\end{abstract}


\end{frontmatter}

\section{Introduction}

In clinical practice, physicians must make several important decisions (interventions, medication, follow-up requirements, etc.) based on all available information. The risk assessment of a given event (i.e., its probability) is one of the most important elements in that decision-making. It is often accomplished through the guide of risk score models. They usually result from simple multivariable regression-based methods which take into account several prognostic factors commonly recorded in the routine medical evaluation. They identify the most important predictors (risk factors) and their relative contribution to the occurrence of the analyzed event, being the final risk a sum of scores attributed to each selected variable \cite{Sullivan2004}.
\par
In recent years, machine learning (ML) methods have been proposed as an alternative to such standard approaches \cite{Tay2015, Chen2020a, Davoodi2018}. Those procedures can model more complex relations between the predictors and the output, usually achieving better predictive performance than the score models \cite{Obermeyer2016}. However, this has an associated cost: such methodologies are often seen by the physicians as a "black-box" with a significant lack of interpretability. This minimizes the confidence of clinicians in the generated models and consequently the trust of the patients as well, discouraging their widespread clinical application \cite{Watson2019}. In the broad context of ML, interpretability is related to the ability of the user to understand the predictions produced by the model and retrieve domain knowledge from it \cite{Murdoch2019, Miller2019}. One of the factors that impact the interpretability of a given model is its complexity in terms of size \cite{10.1145/3236009}. It implies that if a given model is too large, even if every single element can be easily described, the complexity resulting from the combination of several elements may significantly impair its global interpretability. So, models which are a priori interpretable may become hard to explain. For example, a decision tree has its comprehensibility limited by its depth, number of rules, or length of rules condition \cite{10.1145/3236009}.
\par
Even so, these ML approaches present some advantages when compared to score models. They can more easily incorporate new features detected as important ones and be applied to incomplete data sets. Moreover, they often give more importance to the training patients with closer information to the one being evaluated. In other words, while the score models can be seen as a more generalized point system, the ML models are more related to personalized medicine. In fact, score models have been found to perform well at the population level but worse at the individual level \cite{Allen2017, Li2019}.
\par
The state-of-the-art score and ML models focus on the prediction of the occurrence risk of a given event (e.g., death), taking into consideration the overall ability of the model to correctly predict that risk. However, they do not discern if the prediction of a particular patient is trustworthy. Even though some works have been developed about when a prediction should be or not be trusted \cite{Bosnic2008a, Jiang2018, Schulam2019, Kailkhura2019}, this research has received little attention in the decision-making field. Nevertheless, this is very important when a given model's prediction is used to guide critical decisions as in medical diagnosis. This allows the clinical staff to estimate how much they can rely on the algorithm output, identifying when the prediction is more likely to be misleading. It extends the total information available to make the best decisions possible. To the best of our knowledge, only one study was developed to assess the reliability in already existing clinical risk models \cite{Myers2020}.
\par
The forecasting of patients' prognosis in the next days, months or years after an acute coronary syndrome (ACS) event is one of the areas where great efforts of the scientific community have been made to develop tools to assist the health professionals. Cardiovascular disease is the leading cause of mortality worldwide \cite{WorldHealthOrganization2018}, which urges such research. The ACS corresponds to a group of conditions related to low blood flow to the myocardium and it includes unstable angina (UA), non-ST-elevation myocardial infarction (NSTEMI) and ST-elevation myocardial infarction (STEMI). 
\par
In this context, several ACS risk score models have been developed from large clinical trial populations, such as the Platelet glycoprotein IIb/IIIa in Unstable angina: Receptor Suppression Using Integrilin (PURSUIT) \cite{Boersma2000}, the Thrombolysis In Myocardial Infarction (TIMI) \cite{Antman2000a, Morrow2000a}, the Global Use of Strategies to Open Occluded Coronary Arteries in Acute Coronary Syndromes (GUSTO) \cite{L.1995, Armstrong1998a}, and the Global Registry of Acute Coronary Events (GRACE) \cite{Granger2003, Fox2006}. In the last few years, several machine learning-based methodologies were proposed as well \cite{Huang2018,Myers2017,Shouval2017,Kwon2019,Myers2019}. About the assessment of confidence level of an individual mortality risk prediction after an ACS event, to the best of our knowledge, only the aforementioned study \cite{Myers2020} was developed, where the authors evaluated the reliability of the GRACE model predictions. 
\par
Our goal is to develop a new methodology that takes the advantageous features of machine learning methods (good performance), but also incorporates the interpretability level of the clinical score models, in order to create a tool able to predict the risk of any type of event. Furthermore, we intend to use the same approach to estimate the reliability of individual predictions, which is rarely available. Considering such characteristics, we aim to present a support system that not only performs well in terms of accuracy but also inspires trust in physicians through their white-box structure, contributing in a decisive way to support their treatment and follow-up decisions. In section 2 (\textit{Methods}), the several steps of the proposed methodology will be deeply described, and some examples will be provided to help understand the procedure and its properties.
\par
Such an approach is considered and validated in the context of the acute coronary syndrome problem. Therefore, in section 3 (\textit{Results}), the outcomes of the application of the proposed procedure to the prediction of the 30-days mortality risk after an ACS event will be presented, and they will be compared with the performance of state-of-the-art models.
\par
Finally, in sections 4 and 5 (\textit{Discussion} and \textit{Conclusion}), some final remarks will be considered, related to: the advantages and potential of the proposed methodology, the model generated for the examined clinical application, some limitations, and possible further work.

\section{Methods}

In this study, we developed a new approach to help the physicians in the decision-making process related to the forecasting of a given medical condition. Often such risk assessment is related to the risk of death after a given event. Therefore, even it can be used for other end-points, in this work it will be referred to as a predictor of mortality risk.
\par
The procedure proposed in this study can be summarized in three sequential steps: 1) creation of several rules through the dichotomization of meaningful risk factors, 2) training of those rules with a machine learning technique to learn which rules are more suitable to evaluate each patient, 3) computation of the predicted mortality risk and its reliability estimate for each patient using such rules. These methods will be detailed in sections \ref{subsec:methods_creationRules} to \ref{subsec:methods_prediction}.
\par
Further information and methods related to the clinical application and validation of the approach in the acute coronary syndromes scenario (dataset, missing values imputation, data balancing and performance assessment) will be described in sections \ref{subsec:method_acsApplication} and \ref{subsec:method_lastMethods}.

\subsection{Creation of rules, their outputs and acceptance} \label{subsec:methods_creationRules}

The first step is the creation of decision rules. It is carried out by taking important predictors (risk factors) and dichotomizing each one of them. The goal is to define a value that corresponds to the best overall separation threshold between the positive (death) and negative (survival) classes for each one of those variables. Therefore, binary rules are created, signalizing which rules (if any) identify the patient as a potential high-risk one, which is of great interest in the clinical context.
\par 
In order to accomplish that, two different approaches were considered: one based on a moving threshold and another based on centroids. The latter presented a more intelligible and better solution and so it is used. In that procedure, for each feature, two centroids are obtained from the training dataset, each one representing a class: the values of patients who died are used to create one of the centroids and the other is formed using the points corresponding to the patients who survived. 
\par
These two centroids represent then two virtual patients: the positive and negative ones. In this work, each centroid is computed using the mean (Euclidean distance) of the values of the corresponding group. However, if very skewed variables are considered, the use of the median may be advisable for performance and interpretability purposes.
\par
\begin{figure}[H]
\centering
    \includegraphics[width=1\textwidth,keepaspectratio]{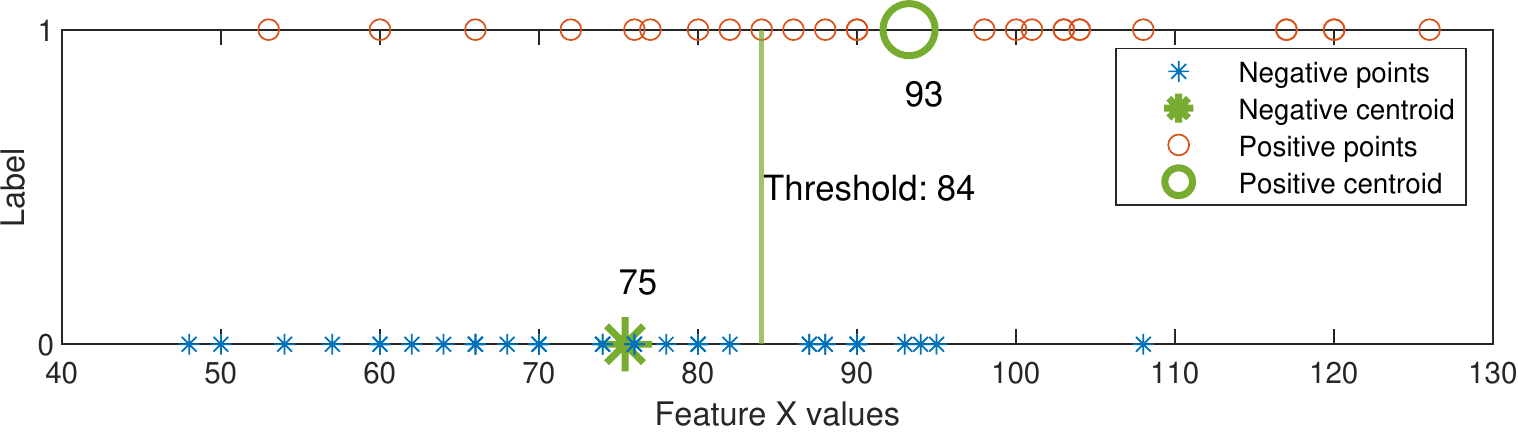}
    \caption{An example of how to obtain the centroids that represent each class, and the corresponding threshold that defines the rule.}
    \label{fig:clusters}
\end{figure}

Having such centroids, the mean distance between them can be seen as the separation threshold. More specifically, if a given sample is closer to the positive centroid then, according to that predictor, the patient is expected to die; otherwise, if it is closer to the negative one, the patient is expected to survive. Thus, the mean distance between centroids is the threshold value that defines the rule for a given feature,  assuring the clinical interpretability of the rule. Figure \ref{fig:clusters} shows an example of this procedure for an arbitrary feature X and its values for the training group of patients.

\begin{figure}[H]
\centering
    \includegraphics[width=1\textwidth,keepaspectratio]{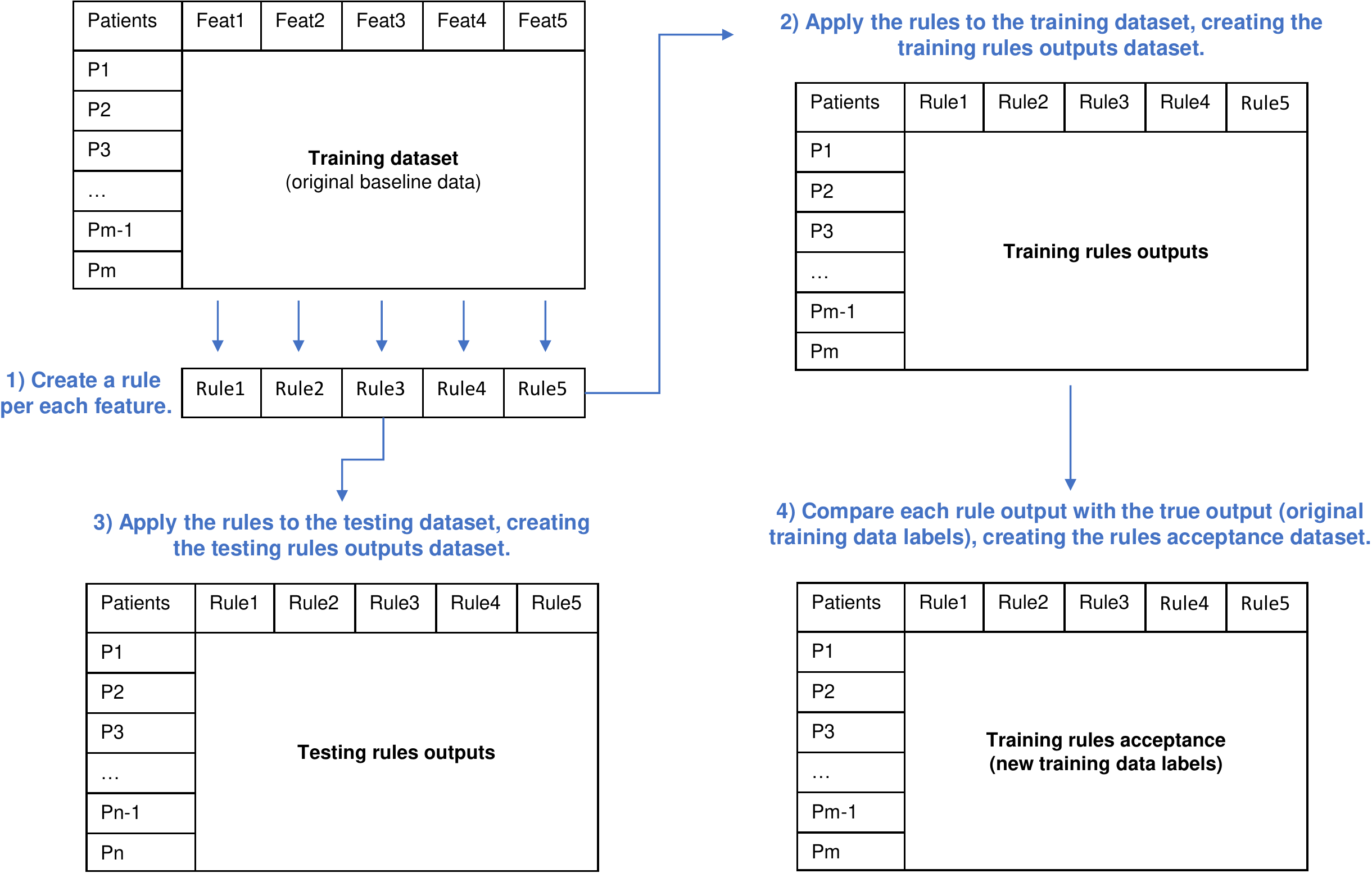}
    \caption{The step-by-step methodology related to the creation of rules and their outputs and training acceptance. Five rules were used to illustrate the method. The \textit{m} corresponds to the number of patients in the training dataset and the \textit{n} to the number of patients in the testing dataset.}
    \label{fig:esquema_pt1}
\end{figure}

For the arbitrary feature X, the rule will then be defined as: \textit{IF value(X)$\geq$84 THEN death; ELSE survival}. These death-or-survival suggestions will be further referred to as  \textbf{\textit{rules outputs}} - figure \ref{fig:esquema_pt1}, step 2. 
\par
Figure \ref{fig:clusters} shows an example of the rule creation for a continuous predictor. Ordinal and binary variables can be directly mapped as well, treating them as continuous ones to obtain the centroids and threshold. The threshold is then used to divide the categories and create the rule. For example, if a given predictor Z has 4 ordinal levels, and the threshold obtained from the centroids is 2.42, and higher the levels higher tendency to death, the rule will be defined as: \textit{IF value(Z)=3 OR value(Z)=4, THEN death; ELSE survival}. For non-ordinal categorical predictors, it is required to first transform them into ordinal or binary ones, for example creating dummy variables through a \textit{one-hot-encoding} approach.
\par
In order to generalize the methodology for all the variables (which have different centroid values), the distance of each point to the centroids was evaluated through the \textbf{\textit{normalized distance}}:

\begin{equation}
    \textrm{normalized distance} =1-\frac{d\_positive}{d\_positive+d\_negative}
\end{equation}

where ${d\_positive}$ is the Euclidean distance of that sample to the positive centroid and the ${d\_negative}$ is the distance to the negative one. This metric is then in the range [0,1]: normalized distance=0 if $d\_negative$=0; normalized distance=1 if $d\_positive$=0. Therefore, the normalized distance being greater than 0.5 is the equivalent to the point being closer to the positive virtual patient, and the rule for any feature Y becomes: \textit{IF normalized distance(Y)$\geq$0.5, THEN death; ELSE survival}. An example of the application of this procedure is showed in table \ref{tab:rules_determination}. Therefore, for each patient, each rule can be positive, when it suggests the patient's death (rule output = 1), or negative, when it suggests the patient's survival (rule output = 0).
\par
That same process is also applied to the testing samples - figure \ref{fig:esquema_pt1}, step 3. 

Once obtained the rules outputs, their acceptance for the training examples is computed. This \textbf{\textit{rule acceptance}} is a binary variable, with a value of 1 if the rule output corresponds to the true label and a value of 0 if they are different - figure \ref{fig:esquema_pt1}, step 4. An example using five arbitrary rules is showed in table  \ref{tab:rules_determination} for a patient who survived. 

\begin{table}[H]
\centering
\scriptsize
\caption{Determination of rules and its acceptance, considering an arbitrary training example in a five rules scenario.}
\label{tab:rules_determination}
\begin{tabular}{c c c c c c}
\hline
 & \textbf{Rule 1} & \textbf{Rule 2} & \textbf{Rule 3} & \textbf{Rule 4} & \textbf{Rule 5} \\ \hline
\textbf{Normalized distance} & 0.32 & 0.12 & 0.73 & 0.64 & 0.20 \\ 
\textbf{Rule output} & 0 & 0 & 1 & 1 & 0 \\
\textbf{True output} & \multicolumn{5}{c}{0 {[}survival{]}} \\ 
\textbf{Rule acceptance} & 1 & 1 & 0 & 0 & 1 \\ \hline
\end{tabular}
\end{table}

\par
Instead of a single binary label related to the death or survival of the patient (the  \textbf{\textit{true output}}), there will be several binary labels, each one related to the acceptance of a rule. Therefore, the rules acceptances are the new data labels.

\subsection{Assessment of rules' acceptance degree for new patients} \label{subsec:methods_acceptancePrediction}

\begin{figure}[H]
\centering
    \includegraphics[width=0.8\textwidth,keepaspectratio]{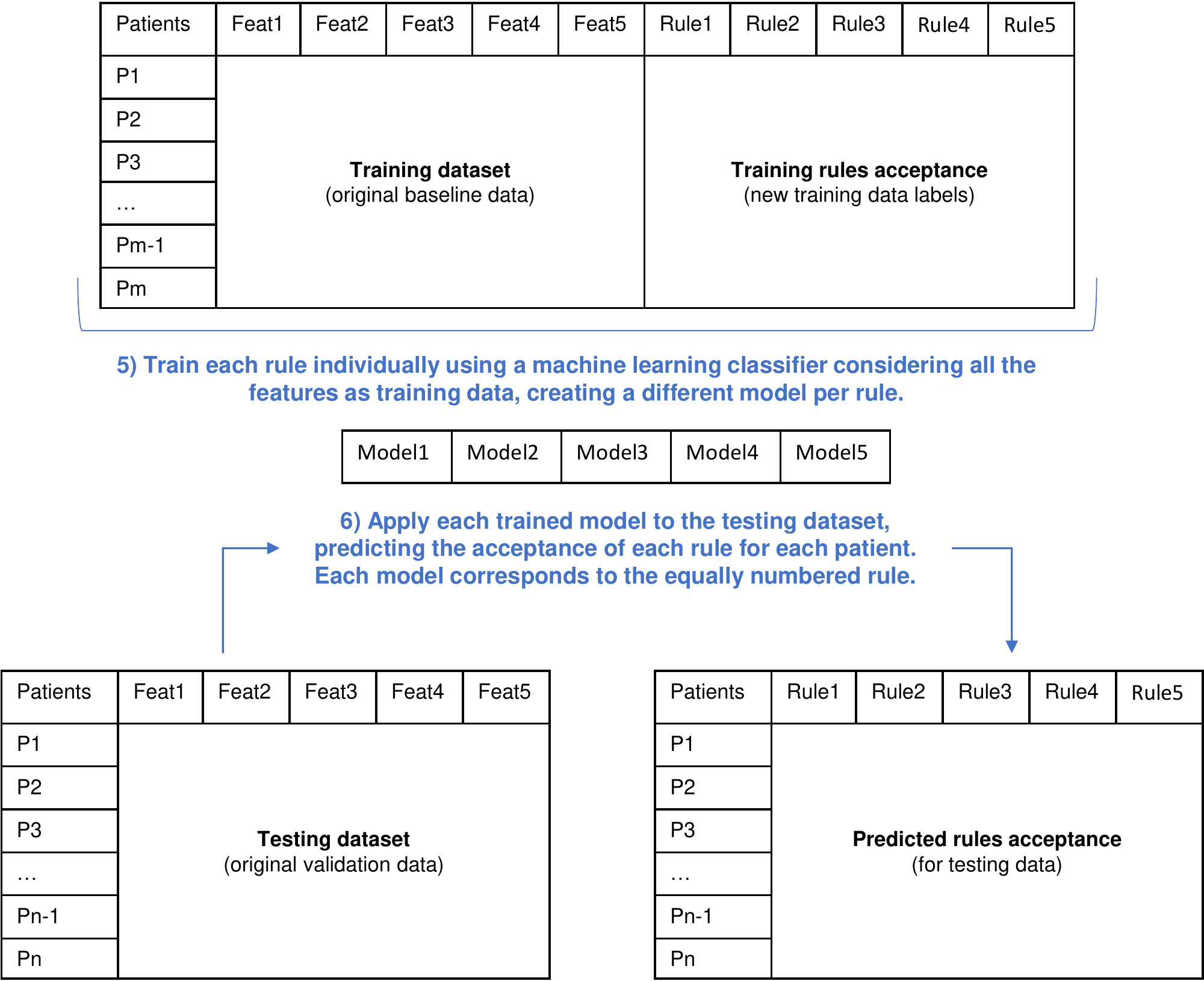}
    \caption{The step-by-step methodology related to the training of rules and the prediction of their acceptance degree in testing data. Five rules were used to illustrate the method. The \textit{m} corresponds to the number of patients in the training dataset and the \textit{n} to the number of patients in the testing dataset.}
    \label{fig:esquema_pt2}
\end{figure}

In the training phase (section \ref{subsec:methods_creationRules}), the rule acceptance of a given rule is obtained by comparing its target (death or survival) with the rule output, and thus only one predictor is required to get it (the risk factor used to create the rule). However, for new samples (simulated by the testing dataset), that target is not known. Therefore, it is necessary to create a model able to predict the rule acceptance for such new samples. In order to perform it, a machine learning procedure is applied, taking all the original values of the risk factors as the training features. The single predictor used to build the rule would not allow a good prediction, and thus all the risk factors are used to train such a model. The idea is then to create a prediction model per rule which estimates if that rule is expected to be suitable to evaluate a given patient - figure \ref{fig:esquema_pt2}, step 5. In fact, all the original feature values are used to train a different model for each rule, i.e., a different model to estimate each new label. In short, as in a standard machine learning procedure, the set of risk factors are used to train a model to predict the label of the new samples, but here such label corresponds to the correctness of a single rule and not to the occurrence of the event (death or survival).
\par
For training patients, the future output (death or survival) is known, and thus the acceptance of each rule is computed as a yes/no response. However, for new patients, it is more meaningful for both algorithms and physicians to have a degree of acceptance, i.e., a likelihood of correctness. Therefore, the prediction models are trained to produce a probability output (predicted probabilities of the rule acceptance) instead of a binary one, and then they are applied to the testing data - figure \ref{fig:esquema_pt2}, step 6. For instance, if the predicted acceptance of a rule is closer to 1 (100\%), there is a high probability that rule is correct for that patient; otherwise, if the predicted acceptance is closer to 0 (0\%), it is likely to be misleading. 
\par
Consequently, only prediction models whose statistical foundation is to produce a probability output (value in the range [0,1]) are used. In our study, a logistic regression model and a neural network with a log-sigmoid activation function in the output layer were considered.
\par
In order to optimize the models for the predictive task where the proposed approach is validated, the parameters of those models were tuned considering their validation performance (AUC value). For the neural network model, the following sets of parameters were considered in a grid search optimization - number of hidden layers: [1,2,3,4], number of neurons in such layers: [2,4,8,16,24,32,64,86,100], and the best combination was selected. Furthermore, regularization procedures were also considered to avoid some possible overfitting. More specifically, elastic-net regularization for the logistic regression model \cite{Zou2005} (with a ratio between L1 and L2 regularization varying in the interval [0,1]) and Bayesian regularization for the neural network model \cite{Burden2008}. These optimization procedures were also applied to the comparison models mentioned in section \ref{subsec:results}: standard logistic regression and neural network methods. For both the proposed approach and comparison models, it was verified that regularization did not statistically improve the validation performance, and so a simpler methodology (no regularization) was considered. Considering the neural network, the performance of the methods did not improve with more than 2 hidden layers and 8 neurons. More specifically, an architecture with 8 neurons in the first layer and 4 neurons in the second one was used for the proposed approach, and an architecture with 8 neurons in both hidden layers was considered for the comparison approach (standard neural network model). 

\subsection{Prediction of mortality risk and reliability estimation} \label{subsec:methods_prediction}

Knowing the outputs of each rule (figure \ref{fig:esquema_pt1}, step 3) and obtaining their predicted acceptance (figure \ref{fig:esquema_pt2}, step 6), the forecasting of each patient is assessed through two computed values: the predicted mortality risk and the predicted reliability estimate. 
\par
To forecast the mortality risk is the major goal in these clinical studies, as it allows the physicians to better identify the patients that are at higher risk and the ones with lower risk, and so defining a suitable treatment for each one. In this work, we intended to create a risk assessment tool considering all the created rules but giving more importance to the ones that are more likely to be correct. Therefore, we start by computing a score based on the mean of the multiplication of the dichotomized outcome of each rule (its binary output) and
the corresponding predicted probability of being corrected (rule acceptances obtained as the output of the models related to each one of the rules):

\begin{equation}
    \textrm{predicted mortality score} = \frac{\sum_{i=1}^{r} (rule\textrm{ } output)_{i} ( predicted  \textrm{ } rule \textrm{ } acceptance)_{i}}{r},
    \label{eq:score}
\end{equation}

where \textit{r} is the total number of rules. Until now, we considered a rule output value of 0 for the rules that suggested the patient's survival (negative rules). However, that value would lead to the removal of all the portions of the equation \ref{eq:score} that are related to the negative rules. Therefore, here, it is considered a rule output of -1 for those rules instead. Thus, the obtained score is a value in the range [-1,1], which is then converted back to the range [0,1]: 
\begin{equation}
    s_{i}= \frac{t_{i}-\text{min}(t)}{\text{max}(t)-\text{min}(t)},
\end{equation}
where $s_{i}$ is the score in the range [0,1], $t_{i}$ is the score in the range [-1,1], $\text{min}(t)$=-1 an $\text{max}(t)$=1.
\par
Ideally, the rules that suggest the correct output will have a higher predicted acceptance and the others a lower one. Therefore, the score is expected to tend to 0 in case of survival or to 1 in case of death. 
\par
In the proposed methodology, new labels (the decision rules) are created and used, instead of mapping directly the features into the outcomes. Thus, a calibration step is required to convert the predicted score into a meaningful predicted risk. Such a procedure is executed by creating a logistic regression model, based on the predicted scores in the training dataset and their corresponding true label (event or no event). The logistic regression calibration model is then applied to the predicted scores to obtain the final predicted risks:

\begin{equation}
    \textrm{predicted mortality risk} = 
    \frac{exp [\beta_0 + \beta_1  (  predicted  \textrm{ }mortality  \textrm{ } score)]}{1 + exp [\beta_0 + \beta_1   (predicted  \textrm{ }mortality  \textrm{ } score)]},
    \label{eq:risk}
\end{equation}

where $\beta_0$ and $\beta_1$ are the coefficients obtained for the calibration model. In this study, we obtained a $\beta_1$=8.0, p-value$<$0.001, which represents a very good ability of the calibration procedure to convert the scores into risks.

Often it is only evaluated how correctly that risk is expected to be for the overall population, i.e., if the predicted risk is well obtained in general. However, usually, there is no way to know the trustfulness of the algorithm for a particular prediction, i.e., to identify patients where that forecasting is expected to be less accurate. The use of calibration curves helps to identify how likely the predictions are to be over or underestimated, based on the relation between the estimated rate of events and the observed one. Even so, this estimation is not often available, as calibration is less assessed than discrimination in research studies \cite{VanCalster2019}. Furthermore, the calibration curves provide a general reliability of predictions at a given risk level (at a given risk X, the reliability is the same for all predictions), but not an individual one (at a given risk X, attribute a personalized and more adjustable reliability to each prediction).
\par
Our goal was to create a tool that is able to provide a degree of confidence for each risk prediction, i.e., its reliability estimation. The goal is to estimate how reliable is the risk prediction for a given patient, providing physicians with additional valuable information: if they can trust the method being used and accept its outcome or, on the other hand, if the model is providing a very uncertain outcome. This has not only the potential to improve the decisions made by the clinical staff but also boost their confidence in this methodology as it becomes more transparent.
\par
If a given patient dies then it is expected that the predicted acceptance degree of the rules with a positive output is much higher than the one of the rules that suggest the negative class; the opposite happens if the patient survives. Considering this evidence, a reliability estimation may be derived. Some approaches were explored, and one was found to be more suitable and to perform better.
\par
More specifically, the higher the difference between the mean predicted acceptance probabilities of the rules that suggest the patient's death and the ones that suggest the patient's survival, the more confidence there is in the predicted risk (more reliable). For example, for a patient who survives, ideally, the predicted acceptance degree of the negative rules would be 100\% and for the positive rules it would be 0\%, so the reliability estimate is 100\%. Thus, the lower the difference between such predicted acceptances, the lower the reliability. 
\par
Therefore, the reliability estimation of a particular prediction was computed as the absolute value of the difference between the means of the predicted acceptances of positive rules (output=1, death) and negative rules (output=0, survival):

\begin{multline}
    \textrm{predicted reliability estimate} = \\
    =  \left |\frac{1}{p} \sum_{j=1}^{p} (predicted  \textrm{ }rule \textrm{ } acceptance)_{j} - \frac{1}{q}\sum_{k=1}^{q} (predicted  \textrm{ }rule \textrm{ } acceptance)_{k} \right | 
    \label{eq:reliab}
\end{multline}

where $p$ is the number of rules that suggest the patient's death and $q$ is the number of rules that suggest the patient's survival, as obtained in section \ref{subsec:methods_creationRules}. 
\par

The equation \ref{eq:score}, related to the mortality score computation, can also be represented as:

\begin{multline}
    \textrm{predicted mortality score} = \\ 
    =  \frac{1}{r}\left [\sum_{j=1}^{p} (predicted  \textrm{ }rule \textrm{ } acceptance)_{j} - \sum_{k=1}^{q} (predicted  \textrm{ }rule \textrm{ } acceptance)_{k}  \right ]  = \\ 
    = \frac{p}{r}\left [ \frac{1}{p} \sum_{j=1}^{p} (predicted  \textrm{ }rule \textrm{ } acceptance)_{j}\right ] - \frac{q}{r}\left [  \frac{1}{q}\sum_{k=1}^{q} (predicted  \textrm{ }rule \textrm{ } acceptance)_{k}  \right ]
    \label{eq:score2}
\end{multline}

Therefore, it is possible to observe that the computation of the mortality score before normalization (equation \ref{eq:score2}), is a weighted version of the computation of the reliability estimate (equation \ref{eq:reliab}). Thus, the latter is a special case (absolute value and weights=1) of the former. Furthermore, comparing both equations, it is possible to observe that it is possible to directly compute the reliability estimate from the mortality score (before normalization) when the weights ($p/r$ and $q/r$) are equal, i.e., when the number of rules that suggest the patient's death ($p$) is equal to the number of rules that suggests patient's survival ($q$). This is only possible to obtain if the total number of rules ($r$) is even. For such particular scenario:

\begin{equation}
    \textrm{predicted reliability estimate} =  \left\lvert 2 (predicted \textrm{ } mortality \textrm{ } score \textrm{ }before \textrm{ }normalization) \right\rvert
\end{equation}

which can also be presented in relation to the predicted mortality score after normalization to the interval [0,1] as:

\begin{equation}
    \textrm{predicted reliability estimate} = \left\lvert  4  (predicted  \textrm{ } mortality  \textrm{ }score  \textrm{ } after  \textrm{ } normalization) - 2 \right\rvert
\end{equation}

However, this is an exceptional case. For every other scenario, it is not possible to directly compute the reliability estimate from the mortality score. Thus, in general, the assessment of the likelihood that a given prediction is correct (reliability estimate) is obtained independently of the value of such prediction (mortality score).

Table \ref{tab:reliability_estimations} shows an example of two patients, the predicted acceptance degree of five arbitrary rules and their output (0 for survival and 1 for death). The predicted mortality score is presented as well. As it is possible to see in that table, the risk of death for patient 1 is higher than the one for patient 2. On the other hand, the reliability in the predicted mortality score of patient 2 is much higher than the one for patient 1, which means there is much more confidence in the score predicted for patient 2.

\begin{table}[H]
\centering
\scriptsize
\caption{Two examples of how to obtain and explain the reliability estimation, and the prediction score.}
\label{tab:reliability_estimations}
\begin{tabular}{c c c c c}
\hline
 & \multicolumn{2}{c}{\textbf{Patient 1}} & \multicolumn{2}{c}{\textbf{Patient 2}} \\ 
\multirow{-2}{*}{} & \textbf{\begin{tabular}[c]{@{}c@{}}predicted rule\\ acceptance\end{tabular}} & \textbf{\begin{tabular}[c]{@{}c@{}}rule\\ output\end{tabular}} & \textbf{\begin{tabular}[c]{@{}c@{}}predicted rule\\ acceptance\end{tabular}} & \textbf{\begin{tabular}[c]{@{}c@{}}rule\\ output\end{tabular}} \\ \hline
\textbf{Rule 1} & 73 \% & 0 & 88 \% & 0 \\ 
\textbf{Rule 2} & 91 \% & 0 & 95 \% & 0 \\ 
\textbf{Rule 3} & 41 \% & 1 & 24 \% & 1 \\ 
\textbf{Rule 4} & 34 \% & 1 & 11 \% & 1 \\ 
\textbf{Rule 5} & 70 \% & 0 & 91 \% & 0 \\ 
\multicolumn{5}{l}{} \\ 
\textbf{\begin{tabular}[c]{@{}c@{}}Predicted mortality\\ score\end{tabular}} & \multicolumn{2}{c}{\textbf{0.34}} & \multicolumn{2}{c}{\textbf{0.26}} \\ 
\multicolumn{5}{l}{} \\ 
\textbf{\begin{tabular}[c]{@{}c@{}}Mean acceptance\\ of positive rules\end{tabular}} & \multicolumn{2}{c}{37.50 \%} & \multicolumn{2}{c}{17.50 \%} \\ 
\textbf{\begin{tabular}[c]{@{}c@{}}Mean acceptance \\ of negative rules\end{tabular}} & \multicolumn{2}{c}{78.00 \%} & \multicolumn{2}{c}{91.33 \%} \\ 
\textbf{\begin{tabular}[c]{@{}c@{}}Predicted reliability\\ estimate\end{tabular}} & \multicolumn{2}{c}{\textbf{40.50 \%}} & \multicolumn{2}{c}{\textbf{73.83 \%}} \\ \hline
\end{tabular}
\end{table}

\subsection{Application of the procedure to the ACS scenario} \label{subsec:method_acsApplication}

The procedure described in sections \ref{subsec:methods_creationRules} to \ref{subsec:methods_prediction} was applied and studied in a real problem related to the risk prediction of the 30-days all-cause mortality after an acute coronary syndrome (ACS) event.
\par 
The research resulted from the analysis of data from two different cohorts, both from Portuguese hospitals, related to patients admitted and diagnosed with an ACS by the physicians according to the guideline criteria. Both cohorts (one with 686 and the other with 590 patients) were unified into a single dataset. From the resulting final cohort, were excluded the patients who did not respect the following criteria: to have at least a follow-up of 30 days from the hospital admission day, to have a follow-up result (death or survival), and to have information about the death day (if applicable). According to these considerations, 1111 patients entered in our analysis.
\par
From the variables available in the initial cohorts, only the ones registered in both cohorts were maintained at a first stage, and the necessary unit conversions were performed. This data collection included several types of information, such as demographics, baseline characteristics, previous medication and the presence of comorbidities. In a second stage, all the variables with more than 15\% missing values were excluded, remaining 28 of them. The 15\% value was chosen as a good cutoff for this selection, observing the missing distributions of values in the dataset. It was verified that the variables most commonly measured and registered in the clinical practice and the ones that are often identified as risk factors for the ACS problem were included in this group (15\% of missing values at most).
\par
The analysis included also patients who received any type of in-hospital reperfusion after admission, such as percutaneous coronary intervention (PCI) or coronary artery bypass surgery (CABG).

\subsection{Performance evaluation, missing imputation and data balancing} \label{subsec:method_lastMethods}

Due to the small number of available samples in the problem being considered, the proposed approach was evaluated in the aforementioned dataset through a Monte Carlo cross-validation (MCCV) with 1000 repetitions to obtain a more realistic and robust evaluation, maintaining the death rate in each train-test sampling (stratification). From all data, 80\% was used to form the training dataset (model development) and 20\% for the testing one (model validation). 
\par
The ability to correctly predict the mortality risk was assessed through the AUC - area under the receiver operating characteristic (ROC) curve, which is commonly referred to as C-statistic in clinical studies. The threshold that stratifies the patients into the low or high-risk groups must reflect the overall clinical context, i.e., to not only be based on the available data but also on management, economic or healthy issues \cite{Wynants2019}. Therefore, the AUC is a suitable metric as it does not consider a particular threshold. However, the AUC alone may not be a suitable measure of the performance, especially in imbalanced scenarios \cite{Zou2016}. 
\par
In this situation, the dichotomization of patients into low or high-risk groups is also an important definition as some treatment guidelines depend on such stratification. So, to evaluate the overall stratification ability of the model, it was also assessed its geometric mean (GM):

\begin{equation}
    \textrm{Geometric mean} = \sqrt{(sensitivity)(specificity)} = \sqrt{\left (  \frac{TP}{TP+FN}\right )  \left ( \frac{TN}{TN+FP} \right ) },
\end{equation}

where \textit{TP}=true positives, \textit{TN}=true negatives, \textit{FP}=false positives, \textit{FN}=false negatives. So, the \textit{sensitivity} is the proportion of patients who died identified as high-risk ones, and \textit{specificity} is the proportion of patients who survived identified as low-risk ones.

Even if the definition of the stratification threshold may require further analysis, it was necessary to select one to compute the TN, TP, FN and FP values. In order to estimate such metrics, in this study, the chosen threshold was the one that maximized the geometric mean in the training dataset.

For a better clinical assessment, two other discrimination metrics were considered, the negative predictive value (NPV) and the positive predictive value (PPV):

\begin{equation}
    \textrm{Negative predictive value} =  \frac{TN}{TN+FN}
\end{equation}

\begin{equation}
    \textrm{Positive predictive value} =  \frac{TP}{TP+FP}
\end{equation}

For the assessment of NPV and PPV, the TN, TP, FN and FP values were computed for a stratification threshold that achieves a high sensitivity, which is usually an important criterion in clinical applications. More specifically, a cut-off that achieves 80\% of sensitivity was considered. 

\par
Besides the discrimination metrics (AUC, GM, NPV and PPV), the risk prediction performance was also assessed through calibration curves. About the reliability of each prediction, the performance of the proposed methodology was analyzed through the correlation between the predicted reliability estimates and the misclassifications rate.
\par
The imputation of values for missing data was performed individually for each one of the MCCV repetitions, using a 10-nearest-neighbor imputation. More specifically, the training and testing missing values were replaced by the mean and mode of the ten training examples with the lowest Euclidean distance to them, for continuous and categorical variables respectively. 
\par
One of the main goals of the proposed methodology is to have meaningful rules, estimating the probability of each one of them to be correct. As a consequence of the imbalanced data, the rules more likely to be correct for the patients who survive would always have a high predicted acceptance, even for the positive patients, while the rules expected to be right for the patients who died would always have a low predicted acceptance. 
\par 
In order to overcome this issue, an undersampling of negative samples in the training dataset was performed. This procedure originated a final 1.5:1 negative-positive ratio. Thus, the general predicted acceptance of the rules who suggested a positive output is higher than the ones who suggested a negative output, for patients who died, and the opposite for patients who survived, as desirable. Fortunately, the smaller size of the dataset used for training did not have a substantial effect on model performance. This undersampling was executed selecting random negative samples, and it was only applied for the machine learning step (creation of the rule acceptance prediction models). In the remaining steps (rules creation,  missing data imputation,  and model calibration) all the available training samples were considered.
\par
Even this necessity of data balancing was verified for this specific problem, it is expected that the same process may be required for the majority of clinical conditions, as typically the number of positive cases (occurrence of the analyzed event) is lower than the number of negative ones (no event).

\section{Results}

\subsection{Patient characteristics}

The overall information about the variables considered in this study is presented in table \ref{tab:baseline_info}. In addition to the 28 pre-treatment variables, reperfusion rates are also reported. The 30-days all-cause death rate is 4.95\%.

\subsection{Feature selection}

The group of variables that present a lower p-value of univariate analysis in our work (mainly the ones with p$<$0.01) is in agreement with the group of variables identified as risk factors in the several ACS studies referred to in the introduction section. As a smaller p-value is associated with a larger statistical incompatibility that the positive and negatives classes have the same distribution \cite{Wasserstein2016}, we considered the selection of variables with p$<$0.01 as a reasonable criterion for the selection of risk factors in this work in a first stage. Furthermore, due to the small size of the dataset, a cutoff of 0.01 also imposes a more rigorous selection, giving more confidence on the selected set of features. Therefore, eleven variables were primarily selected: diagnosis, age, antecedent of stroke/transient ischemic attack, systolic and diastolic blood pressures, heart rate, Killip class, left ventricular ejection fraction, elevation of cardiac markers, glucose, hemoglobin.
\par
From those, two variables were discarded: diastolic blood pressure (DBP) and elevation of cardiac markers, mainly due to their correlation with other predictors. The information of the latter is already implied in the diagnosis because it is used to discriminate between myocardial infarction and unstable angina. The DBP is known to have a strong correlation with the systolic blood pressure (SBP) \cite{Gavish2008}, which was verified in this study (r$\approx$0.7); clinical studies only considered the SBP and, therefore, the same approach was followed here. 
\par
We are dealing with a small dataset and a small number of events, and a high rate of variables per event can lead to overfitting of the generated model. Therefore, we performed a final feature selection. From those nine variables, we selected only the ones that are used in at least one of the four risk score models most used in the clinical practice (GRACE, TIMI, PURSUIT, GUSTO). This is a conservative and robust selection because the clinical relevance of those risk factors in this context was already been validated in large datasets. Moreover, this induces more confidence in clinical professionals as well. Thus, the final group of features to be used in our approach is composed of six variables: diagnosis, age, systolic blood pressure, heart rate, Killip class, and previous stroke/TIA.

\begin{table}[H]
\centering
\scriptsize
\caption{General information of the patients used in the analysis of this study. For the binary variables - diabetes, smoking, hypertension, stroke/TIA, myocardial infarction, unstable angina, PTCA, CABG, ST-segment deviation, elevation of cardiac markers, all types of medication, and reperfusion - the values are related to the positive class: the existence of such condition or characteristic. The reported values are related to the median and interquartile range for the continuous variables, and to the frequencies and percentages for the categorical ones.}
\label{tab:baseline_info}
\begin{tabular}{lccccc}
\hline
\multicolumn{2}{c}{} & \textbf{Total} & \textbf{Survival} & \textbf{Death} & \textbf{p-value}\\ \hline
\multicolumn{2}{l}{No. of patients} & 1111 & 1056 & 55 &  \\
\multicolumn{1}{c}{} &  &  &  &  &  \\
Diagnosis & UA & 247 (22.2\%) & 244 (23.1\%) & 3 (5.5\%) & \textless0.001\\
 & NSTEMI & 452 (40.7\%) & 433 (41.0\%) & 19 (34.5\%) &  \\
 & STEMI & 402 (36.2\%) & 369 (34.9\%) & 33 (60.0\%) &  \\
\multicolumn{6}{l}{\textbf{Demographics}} \\
Gender & Female & 260 (23.4\%) & 240 (22.7\%) & 20 (36.4\%) & 0.020\\
 & Male & 851 (76.6\%) & 816 (77.3\%) & 35 (63.6\%) &  \\
\multicolumn{2}{l}{Age} & 64 (54-73) & 64 (54-73) & 72 (63-81) & \textless 0.001 \\
\multicolumn{6}{l}{\textbf{Risk factors / comorbodities}} \\
\multicolumn{2}{l}{Diabetes} & 280 (25\%) & 261 (24.7\%) & 19 (34.5\%) & 0.106 \\
\multicolumn{2}{l}{Smoking - current smoker} & 288 (25.9\%) & 279 (26.4\%) & 9 (16.4\%) & 0.094 \\
\multicolumn{2}{l}{Smoking - former smoker} & 268 (24.1\%) & 261 (24.7\%) & 7 (12.7\%) & 0.038 \\
\multicolumn{2}{l}{Hypertension} & 695 (62.6\%) & 658 (62.3\%) & 37 (67.3\%) & 0.485 \\
\multicolumn{2}{l}{PAD} & 76 (6.8\%) & 70 (6.6\%) & 6 (10.9\%) & 0.207 \\
\multicolumn{6}{l}{\textbf{Clinical antecedents}} \\
\multicolumn{2}{l}{Stroke and/or TIA} & 85 (7.7\%) & 71 (6.7\%) & 14 (25.5\%) & \textless 0.001 \\
\multicolumn{2}{l}{Myocardial infarction} & 327 (29.4\%) & 314 (29.7\%) & 13 (23.6\%) & 0.367 \\
\multicolumn{2}{l}{Unstable angina} & 163 (14.7\%) & 159 (15.1\%) & 4 (7.3\%) & 0.120 \\
\multicolumn{2}{l}{PTCA} & 233 (21.0\%) & 227 (21.5\%) & 6 (10.9\%) & 0.067 \\
\multicolumn{2}{l}{CABG} & 128 (11.5\%) & 126 (11.9\%) & 2 (3.6\%) & 0.064 \\
\multicolumn{6}{l}{\textbf{Characteristics at presentation}} \\
\multicolumn{2}{l}{SBP (mmHg)} & 140 (121-160) & 140 (121-160) & 130 (107-150) & \textless 0.001 \\
\multicolumn{2}{l}{DBP (mmHg)} & 80 (70-90) & 80 (70-90) & 70 (63-82) & \textless 0.001 \\
\multicolumn{2}{l}{Heart rate (bpm)} & 75 (64-88) & 75 (64-87) & 90 (76-25) & \textless 0.001 \\
Killip class & I & 915 (82.4\%) & 888 (84.1\%) & 27 (49.1\%) & \textless 0.001\\
 & II & 103 (9.3\%) & 92 (8.7\%) & 11 (20.0\%) &  \\
 & III & 65 (5.9\%) & 54 (5.1\%) & 11 (20.0\%) &  \\
 & IV & 10 (1.0\%) & 6 (0.6\%) & 4 (7.3\%) &  \\
LEVF & \textgreater{}50 \% & 677 (60.9\%) & 664 (62.9\%) & 13 (23.6\%) & \textless 0.001\\
 & {[}30,50{]} \% & 293 (26.4\%) & 278 (26.3\%) & 15 (27.3\%) &  \\
 & \textless{}30 \% & 80 (7.2\%) & 70 (6.6\%) & 10 (18.2\%) &  \\
\multicolumn{2}{l}{ST segment deviation} & 745 (67.1\%) & 700 (66.3\%) & 45 (81.8\%) & 0.013 \\
\multicolumn{6}{l}{\textbf{Laboratory results}} \\
\multicolumn{2}{l}{Elevation of cardiac markers\textsuperscript{a}} & 614 (55.3\%) & 572 (54.2\%) & 42 (76.4\%) & 0.001 \\
\multicolumn{2}{l}{Creatinine (mg/dL)} & 1.0 (0.9-1.2) & 1.0 (0.9-1.2) & 1.1 (1.0-1.5) & 0.066 \\
\multicolumn{2}{l}{Glucose (mg/dL)} & 126 (101-169) & 125 (101-167) & 144 (123-209) & \textless{}0.001 \\
\multicolumn{2}{l}{Hemoglobin (g/dL)} & 14.0 (12.8-15.0) & 14.0 (12.9-15.0) & 12.8 (11.8-14.4) & 0.001 \\
\multicolumn{6}{l}{\textbf{Medication pre-admission}} \\
\multicolumn{2}{l}{Antiplatelet drugs} & 526 (47.3\%) & 507 (48.0\%) & 19 (34.5\%) & 0.094 \\
\multicolumn{2}{l}{Beta-blockers} & 292 (26.3\%) & 279 (26.4\%) & 13 (23.6\%) & 0.715 \\
\multicolumn{2}{l}{Calcium channel blockers} & 245 (22.1\%) & 234 (22.2\%) & 11 (20.0\%) & 0.767 \\
\multicolumn{2}{l}{Statin} & 254 (22.9\%) & 242 (22.9\%) & 12 (21.8\%) & 0.918 \\
\multicolumn{2}{l}{Antihypertensive drugs} & 325 (29.3\%) & 307 (29.1\%) & 18 (32.7\%) & 0.485 \\
\multicolumn{1}{c}{} &  &  &  &  &  \\
\multicolumn{2}{l}{Reperfusion} & 621 (55.9\%) & 596 (56.4\%) & 25 (45.5\%) & 0.110 \\ \hline
\multicolumn{6}{p{\dimexpr\linewidth-2\tabcolsep-2\arrayrulewidth}}{UA: unstable angina, NSTEMI: non-ST-elevation myocardial infarction, STEMI: ST-elevation myocardial infarction, PAD: peripheral artery disease, TIA: transient ischemic attack, PTCA: percutaneous transluminal coronary angioplasty, CABG: coronary artery bypass grafting, SBP: systolic blood pressure, DBP: diastolic blood pressure, LVEF: left ventricular ejection fraction. 
\par \textsuperscript{a}The elevation of cardiac markers is related to the elevation above critical thresholds of cardiac injury biomarkers, such as troponin, myoglobin and creatine kinase (CK)-MB, accordingly to clinical guidelines.}
\end{tabular}
\end{table}

\subsection{Prediction of the mortality risk} \label{subsec:results}

As previously mentioned, for the rules acceptance prediction models, the neural network (with a simple 3-layers architecture) and the logistic regression models were considered. No statistical significance (p$>$0.1) was verified in the results obtained for both AUC and geometric mean using the two methods. Thus, the results further presented for the proposed approach were obtained using artificial neural networks, but they represent both methods, i.e., they characterize the proposed methodology in general. The methodology was also compared with three other procedures that are state-of-the-art methodologies: a clinical risk score model based on a system of points, a logistic regression, and an artificial neural network.
\par
The risk prediction in clinical practice is typically performed using models based on a system of points (score models). Such a decision support system is the most easy-to-use and interpretable approach to forecast the occurrence of a clinical event. In this work, we considered the GRACE risk model \cite{Granger2003}, which was developed from a large cohort of patients admitted with acute coronary syndrome events. From the most known ACS risk score models, the GRACE showed to be the one that best predicts the 30-days mortality in a Portuguese study \cite{DeAraujoGoncalves2005}, being also the most used model.
\par
Apart from that, we also applied two standard statistical/machine learning (ML) methodologies, i.e., models directly linking the original feature values with the binary label corresponding to death or survival after 30-days from admission.
\par
Even some procedures have been proposed to promote interpretable clinical risk predictions \cite{Billiet2018, Caicedo-Torres2019, Chen2020, Caruana2015, Gu2020}, such approaches lack further validation beyond their studies. Moreover, often interpretability-based ML methodologies focus on finding the importance attributed to each variable by more complex prediction models and thus retrieve some explainability from it. However, it can be assessed with a simple logistic regression as well. Therefore, from the well-established machine learning methodologies, the logistic regression is still the most interpretable one to deal with risk prediction. Furthermore, recent surveys show that the majority of clinical decision support systems used in emergency departments \cite{Fernandes2020} and in hospital readmission prediction \cite{Artetxe2018} are based on logistic regression, which means it is still the ‘gold standard’.
\par
As we considered an artificial neural network (ANN) as a predictor to estimate the probability of each rule to be correct for each patient, we also applied a ‘pure’ ANN to further compare our approach. As for the standard logistic regression, to establish such comparison, we replaced the rule-based section with a neural network that directly estimates the outcome (event/no event) risk. \par
The three comparison models were applied to the same patients and following a similar procedure as before, i.e., the same methods for missing imputation and performance assessment; no data balancing was performed.
\par
\begin{figure}[H]
\centering
\begin{minipage}{.48\textwidth}
  \centering
  \includegraphics[width=\linewidth]{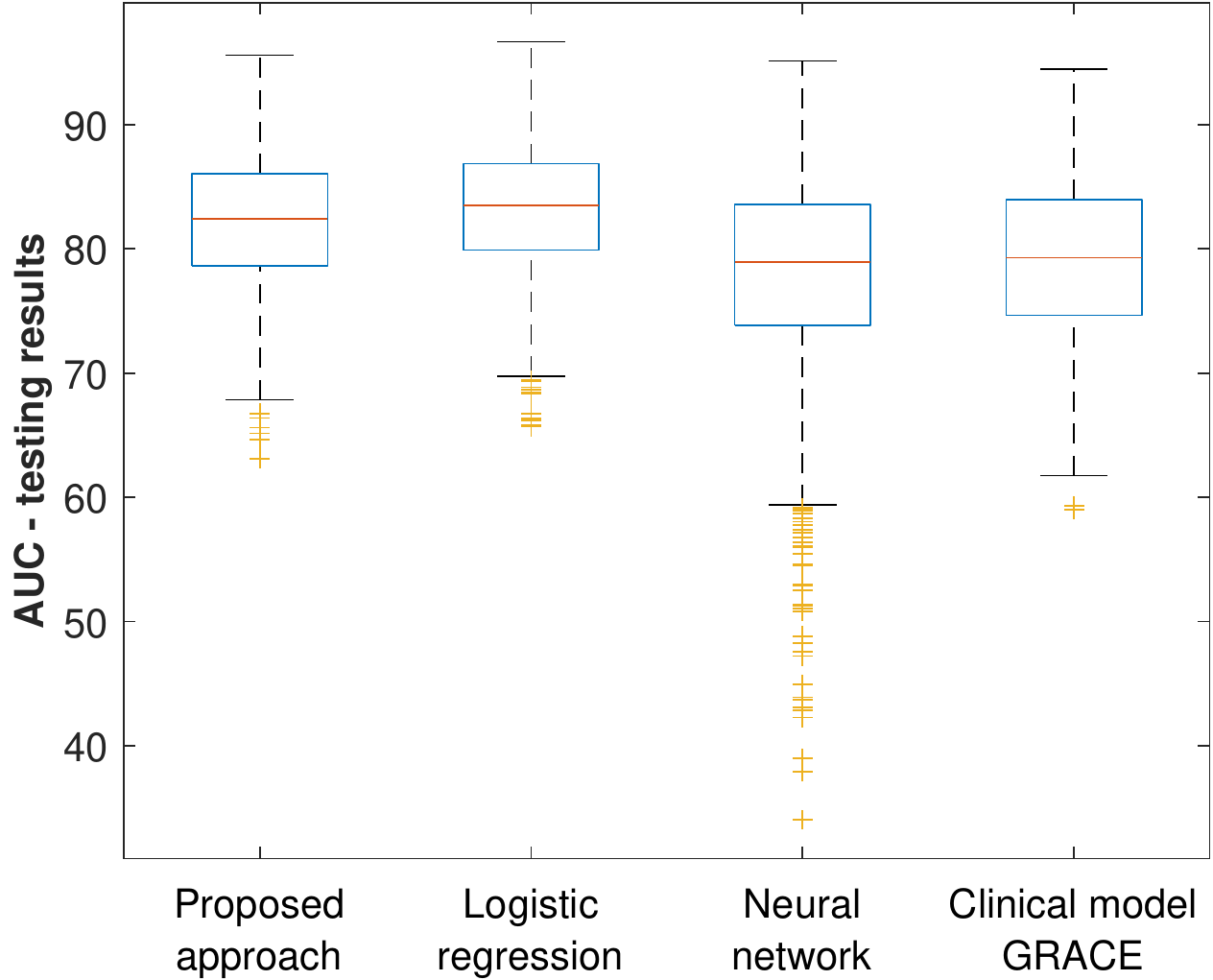}
  \captionof{figure}{Validation performance for the proposed approach and the comparison mododels, according to the area under the ROC curve (AUC) metric. It represents the boxplot of the 1000 runs of Monte-Carlo cross-validation.}
  \label{fig:auc_results}
\end{minipage}\hfill
\begin{minipage}{.48\textwidth}
  \centering
  \includegraphics[width=\linewidth]{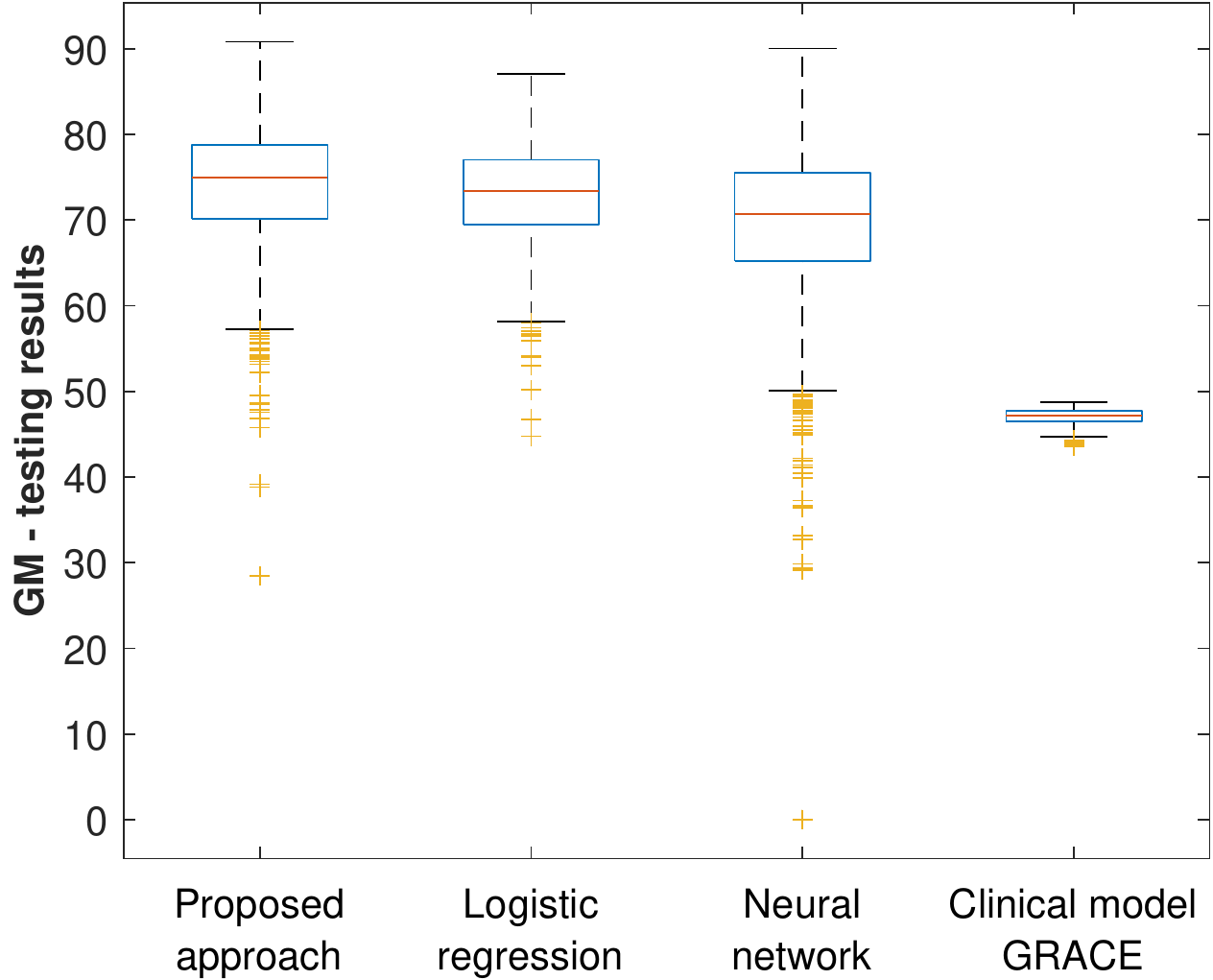}
  \captionof{figure}{Validation performance for the proposed approach and the comparison models, according to the geometric mean (GM) metric.  It represents the boxplot of the 1000 runs of Monte-Carlo cross-validation.}
  \label{fig:gm_results}
\end{minipage}
\end{figure}

The results of the 30-days mortality risk prediction using the proposed approach and the comparison models are analyzed in terms of discrimination and calibration. About the former, the AUC, GM, PPV and NPV results, over the 1000 runs of Monte-Carlo cross-validation, are presented in table \ref{tab:results_mortality}. The associated boxplots related to AUC and GM of testing data are presented in figures \ref{fig:auc_results} and \ref{fig:gm_results}. As explained in the Methods section, the NPV and PPV values are obtained using a fixed cut-off that achieves a sensitivity of 80\%, for all the presented models. The GM is obtained using a cut-off threshold based on the maximization of of the training dataset GM for the proposed approach and the standard machine-learning models, and for the GRACE model it is obtained considering the cut-off suggested in its development study \cite{Granger2003}. 
\par
\begin{figure}[H]
\centering
\begin{minipage}{.48\textwidth}
  \centering
  \includegraphics[width=\linewidth]{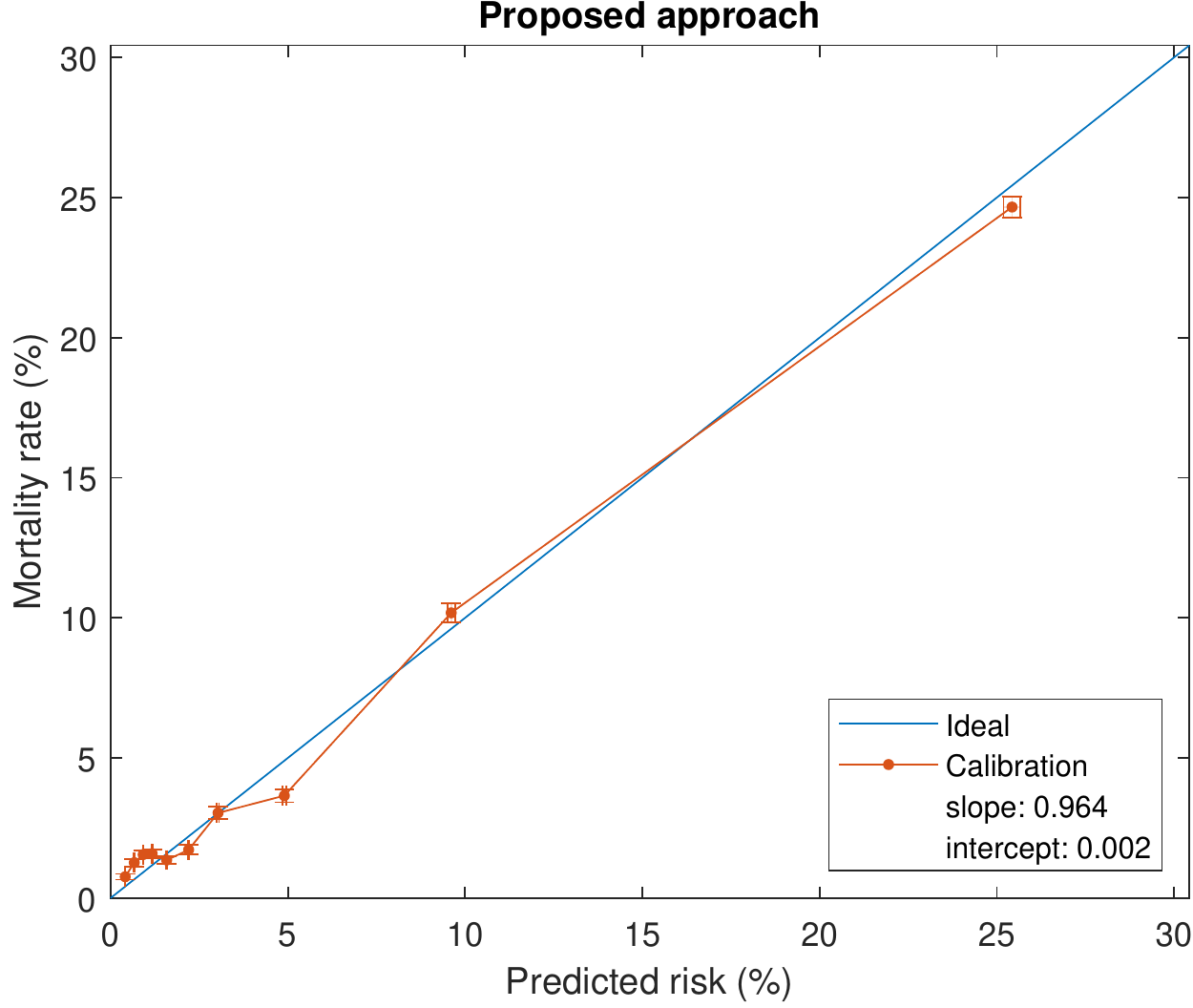}
  \captionof{figure}{Calibration curve for the proposed approach model in the testing dataset. Each point represents the mean and 95\% confidence intervals of the means of the deciles of the predicted risk values, over the 1000 runs of Monte-Carlo cross-validation.}
  \label{fig:proposed_calibration}
\end{minipage}\hfill
\begin{minipage}{.48\textwidth}
  \centering
  \includegraphics[width=\linewidth]{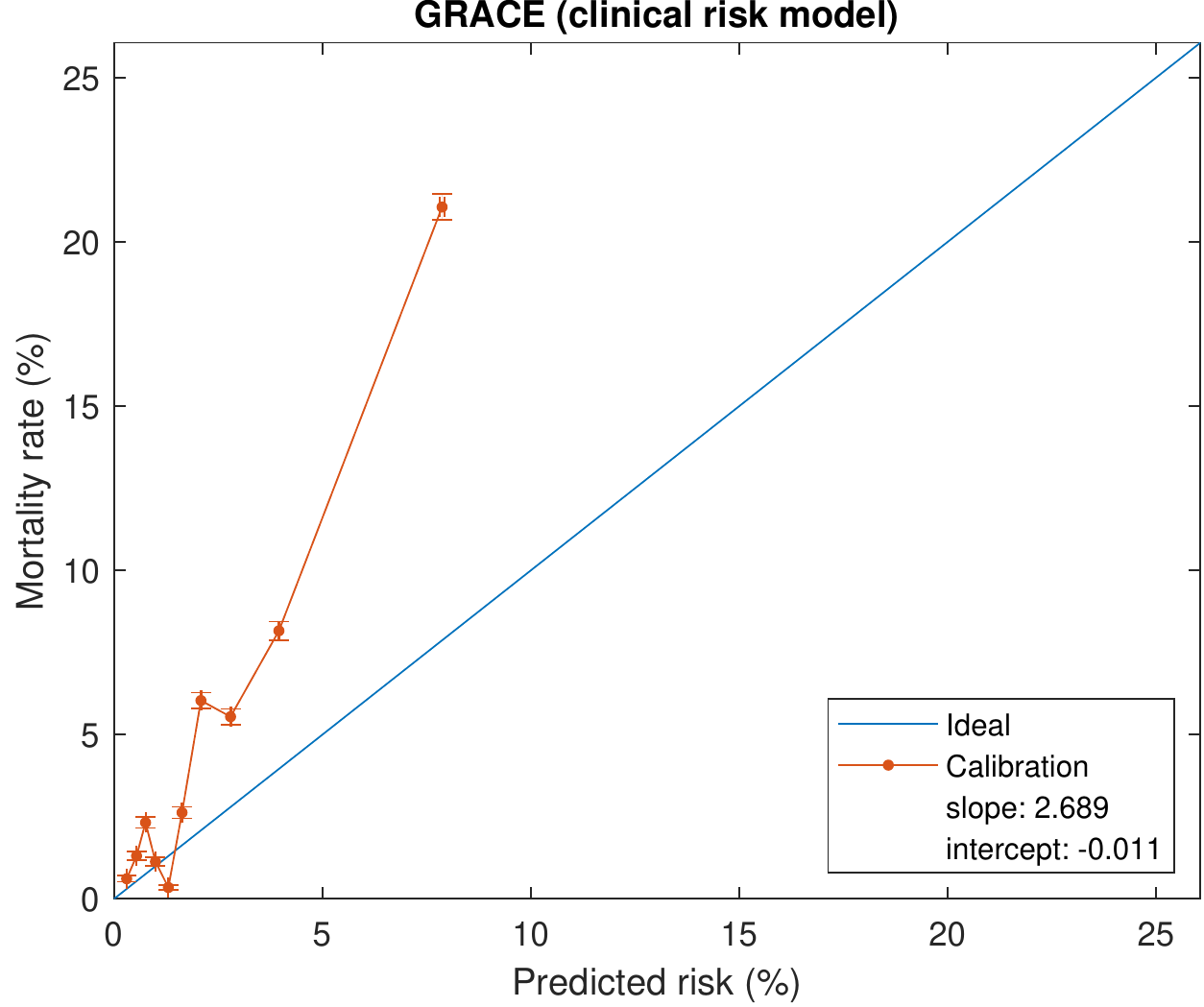}
  \captionof{figure}{Calibration curve for the GRACE risk model in the testing dataset. Each point represents the mean and 95\% confidence intervals of the means of the deciles of the predicted risk values, over the 1000 runs of Monte-Carlo cross-validation.}
  \label{fig:grace_calibration}
\end{minipage}
\end{figure}
\begin{figure}[H]
\centering
\begin{minipage}{.48\textwidth}
  \centering
  \includegraphics[width=\linewidth]{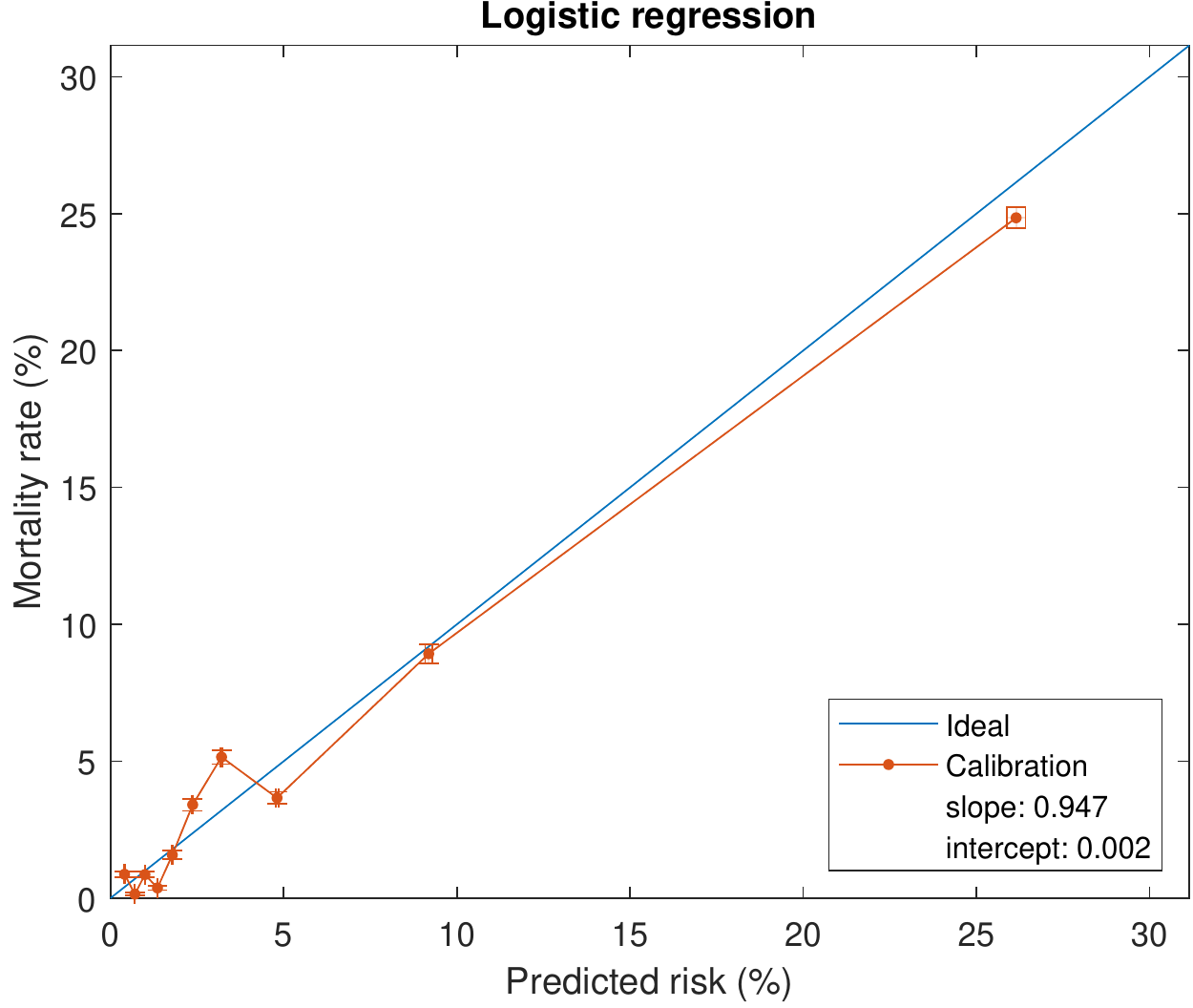}
  \captionof{figure}{Calibration curve for the logistic regression model in the testing dataset. Each point represents the mean and 95\% confidence intervals of the means of the deciles of predicted risk values, over the 1000 runs of Monte-Carlo cross-validation.}
  \label{fig:lr_calibration}
\end{minipage}\hfill
\begin{minipage}{.48\textwidth}
  \centering
  \includegraphics[width=\linewidth]{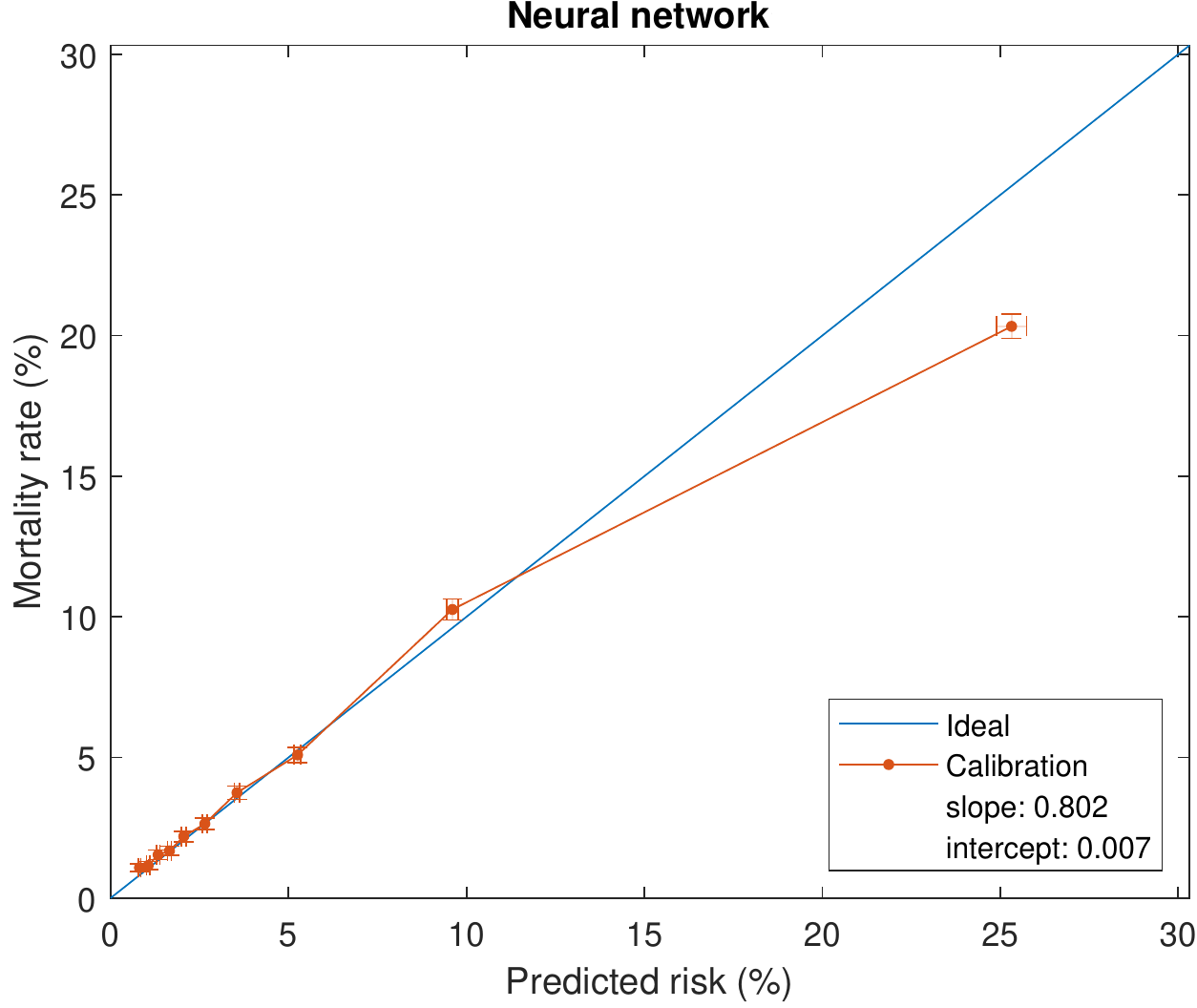}
  \captionof{figure}{Calibration curve for the neural network model in the testing dataset. Each point represents the mean and 95\% confidence intervals of the means of the deciles of the predicted risk values, over the 1000 runs of Monte-Carlo cross-validation.}
  \label{fig:ann_calibration}
\end{minipage}
\end{figure}

As the models are all developed (except the GRACE one) and validated on the same datasets, table \ref{tab:results_deltas} presents the results of the differences for each of the 1000 runs between the proposed method and each one of the comparison models, for all those discrimination metrics. The calibration is then assessed in figures \ref{fig:proposed_calibration} to \ref{fig:ann_calibration}, which show how the mortality rate varies with the predicted mortality risk in the validation group, for the proposed approach and the three comparison models (the standard machine learning methods and the GRACE model).

\begin{table}[H]
\centering
\addtolength{\leftskip} {-2cm}
\addtolength{\rightskip}{-2cm}
\scriptsize
\caption{Obtained results for the prediction of the mortality risk, related to the 1000 runs of Monte Carlo cross-validation (MCCV). They are represented by the 95\% confidence interval. The parameters of the GRACE model are retrieved directly from its development study \cite{Granger2003}, and so a training phase was not performed. The GRACE results are related to its validation in the 1000 testing sets, as done for the other models.}
\label{tab:results_mortality}
\begin{tabular}{ccccccccc}
\hline
\multirow{2}{*}{\textbf{\begin{tabular}[c]{@{}c@{}}Metric\\ (\%)\end{tabular}}} &
  \multicolumn{2}{c}{\textbf{\begin{tabular}[c]{@{}c@{}}Proposed\\ approach\end{tabular}}} &
  \multicolumn{2}{c}{\textbf{\begin{tabular}[c]{@{}c@{}}Logistic\\ regression\end{tabular}}} &
  \multicolumn{2}{c}{\textbf{\begin{tabular}[c]{@{}c@{}}Neural\\ network\end{tabular}}} &
  \multicolumn{1}{c}{\textbf{\begin{tabular}[c]{@{}c@{}}Clinical model\\ GRACE\end{tabular}}} \\
             & Train. & Test. & Train. & Test. & Train. & Test. &  Test. \\ \hline
\textbf{AUC} & [84.1, 84.3]    &  [80.9, 81.7]  & [84.9, 85.1]     & [82.5, 83.3]    & [82.2, 82.8]    & [77.8, 79.0]  & [78.9, 79.6]   \\
\textbf{GM}  & [79.2, 79.4]     &  [73.5, 74.5]   & [77.2, 77.4]    & [72.1, 72.9]    & [76.4, 77.0]    & [69.9, 70.9]     & [47.4, 47.5] \textsuperscript{a}   \\
\textbf{PPV}  & [14.3, 14.6]     &  [16.1, 17.2]   & [13.3, 13.4]    & [15.8, 16.9]    & [13.1, 13.6]    & [13.0, 13.8]    &  [12.6, 13.2] \\
\textbf{NPV}  & [98.5, 98.6]     &  [98.5, 98.6]   & [98.5, 98.5]    & [98.6, 98.7]    & [98.2, 98.4]    & [97.8, 98.3]     &  [98.4, 98.5] \\
\hline
\multicolumn{7}{p{\dimexpr\linewidth-2\tabcolsep-2\arrayrulewidth}}{\footnotesize{Train.: training (development group), Test.: testing (validation group)  \par \textsuperscript{a} In order to compute the geometric mean, the intermediate-risk group was joined to the low-risk one.}}
\end{tabular}
\end{table}

\begin{table}[H]
\centering
\scriptsize
\caption{Differences between the proposed method and the traditional logistic regression and artificial neural network models, for each of the 1000 Monte-Carlo cross-validation runs. The results are represented by the 95\% confidence interval of each difference. The parameters of the GRACE model are retrieved directly from its development study \cite{Granger2003}, and so a training phase was not performed. The GRACE results are related to its validation in the 1000 testing sets, as done for the other models.}
\label{tab:results_deltas}
\begin{tabular}{ccccccc}
\hline
\multirow{2}{*}{\textbf{\begin{tabular}[c]{@{}c@{}}Metric\\ (\%)\end{tabular}}} &
  \multicolumn{2}{c}{\textbf{\begin{tabular}[c]{@{}c@{}}Proposed approach\\ and logistic regression\end{tabular}}} &
  \multicolumn{2}{c}{\textbf{\begin{tabular}[c]{@{}c@{}}Proposed approach\\ and neural network\end{tabular}}} &
  \multicolumn{1}{c}{\textbf{\begin{tabular}[c]{@{}c@{}}Proposed approach\\ and GRACE\end{tabular}}} \\
                                     & Training    & Testing          & Training    & Testing     & Testing\\ \hline
\textbf{$\delta$AUC}    &   [-0.8, -0.6]    & [-1.9, -1.5]   &      [1.5, 2.1]    & [2.5, 3.3] & [1.5, 2.4] \\
\textbf{$\delta$GM}    &    [1.9, 2.1]   & [1.2, 2.0]    &    [2.3, 2.9]      & [3.0, 4.2] & [26.0, 26.9]  \\
\textbf{$\delta$PPV}    &    [1.0, 1.3]   & [-0.1, 0.6]    &    [1.0, 1.3]      & [2.8, 3.7]  & [3.0, 4.5] \\
\textbf{$\delta$NPV}    &    [$<$0.1, $<$0.1]   & [-0.1, -0.1]    &    [0.2, 0.4]      & [0.3, 0.8]   & [0.1, 0.2]  \\ \hline
\end{tabular}
\end{table}

\par
Considering the obtained results, the proposed procedure showed to perform very well in predicting the mortality risk after an ACS event in terms of discrimination metrics (tables \ref{tab:results_mortality} and \ref{tab:results_deltas}, and figures \ref{fig:auc_results} and \ref{fig:gm_results}). Even if it uses a smaller part of the dataset for the prediction models training due to the balancing step, it presents a similar performance when compared to the standard logistic regression: it seems to perform slightly better in terms of GM and slightly worst in terms of AUC. Also, it performs far better than the standard artificial neural network (ANN). The wide confidence intervals of figures \ref{fig:auc_results} and \ref{fig:gm_results} were already expected due to the sample size, which implies a higher variance in each one of the 1000 runs' data-splits. 
\par
The testing calibration curve of the proposed methodology (figure \ref{fig:proposed_calibration}) also suggests a very good generalization ability of the proposed model, with a curve very close to the ideal one, and a consequent slope close to 1. It seems to present a slightly better testing calibration than the standard logistic regression model (figure \ref{fig:lr_calibration}) and the standard ANN model (figure \ref{fig:ann_calibration}). 
\par
The developed approach has also a considerably better performance than the GRACE score model. The lower ability of GRACE to predict the mortality risk, evaluated by the AUC metric, may be explained by two reasons: 1) it has indeed a worse prediction ability than the proposed model, or 2) it is developed from a more heterogeneous population while the suggested method is derived from a Portuguese population where both methods are being validated. The high underfitting verified in the calibration curve (figure \ref{fig:grace_calibration}) supports the idea of the lower performance ability of the GRACE risk model in the dataset being analyzed. 
\par
In relation to the geometric mean, it can not be directly computed for the GRACE model, as that method categorizes the patients into three categories: low, intermediate or high-risk. Therefore, it was necessary to create only two groups, joining the intermediate to another one. Several validation studies of the GRACE model have shown that the death rate of the intermediate group is much closer to the low-risk group than to the high-risk one \cite{Barra2012, Elbarouni2009, Yan2007}, which was also verified in our data. Furthermore, it has been suggested that early intervention seems to be important to prevent future adverse events in patients identified as high-risk by the GRACE model, but not for the patients at low and intermediate-risk \cite{Mehta2009}. Considering such evidence, the following stratification of the GRACE results was performed: low-and-intermediate-risk or high-risk. Even some negative bias may result from such consideration, there is a very relevant difference between the obtained GM values, which suggests that the proposed procedure has a better capability of stratification.
\par
Finally, the NPV and PPV metrics are highly dependent on the prevalence of the dataset \cite{VanStralen2009, Mandrekar2010}; in this case, the rate of patients who died. As this prevalence is very low in this scenario (4.95\%), the NPV is expected to be very high, and the PPV is expected to be low \cite{VanStralen2009, Mandrekar2010}, which is verified in the obtained results. As a consequence of that, it is also expected to obtain a lower variance in the NPV values between the models. Such issue is also verified in the results, as the obtained NPV values are in general equivalent for all the four models: the proposed approach has a NPV performance equivalent to the logistic regression and the GRACE models, and very slightly better than the neural network model. About the PPV values, the proposed approach performs significantly better than the neural network and GRACE models, and very slightly better than the logistic regression one. Thus, those results also support the idea that, besides its intrinsic characteristics which are potentially beneficial for the clinical domain, the methodology developed in this study achieves a good predictive performance.
\par
The suggested approach was repeated without imputation, excluding the patients with missing data in the 6 used features. The results presented a similar training and testing performance (for the validation set, AUC=81\%, GM=73\%, NPV=98\%, PPV=16\%).  Overall, the results are in favor of the effectiveness of the applied missing data imputation procedure.

\subsection{Prediction of the reliability estimate}

\begin{figure}[H]
   \centering
  \includegraphics[width=0.75\linewidth]{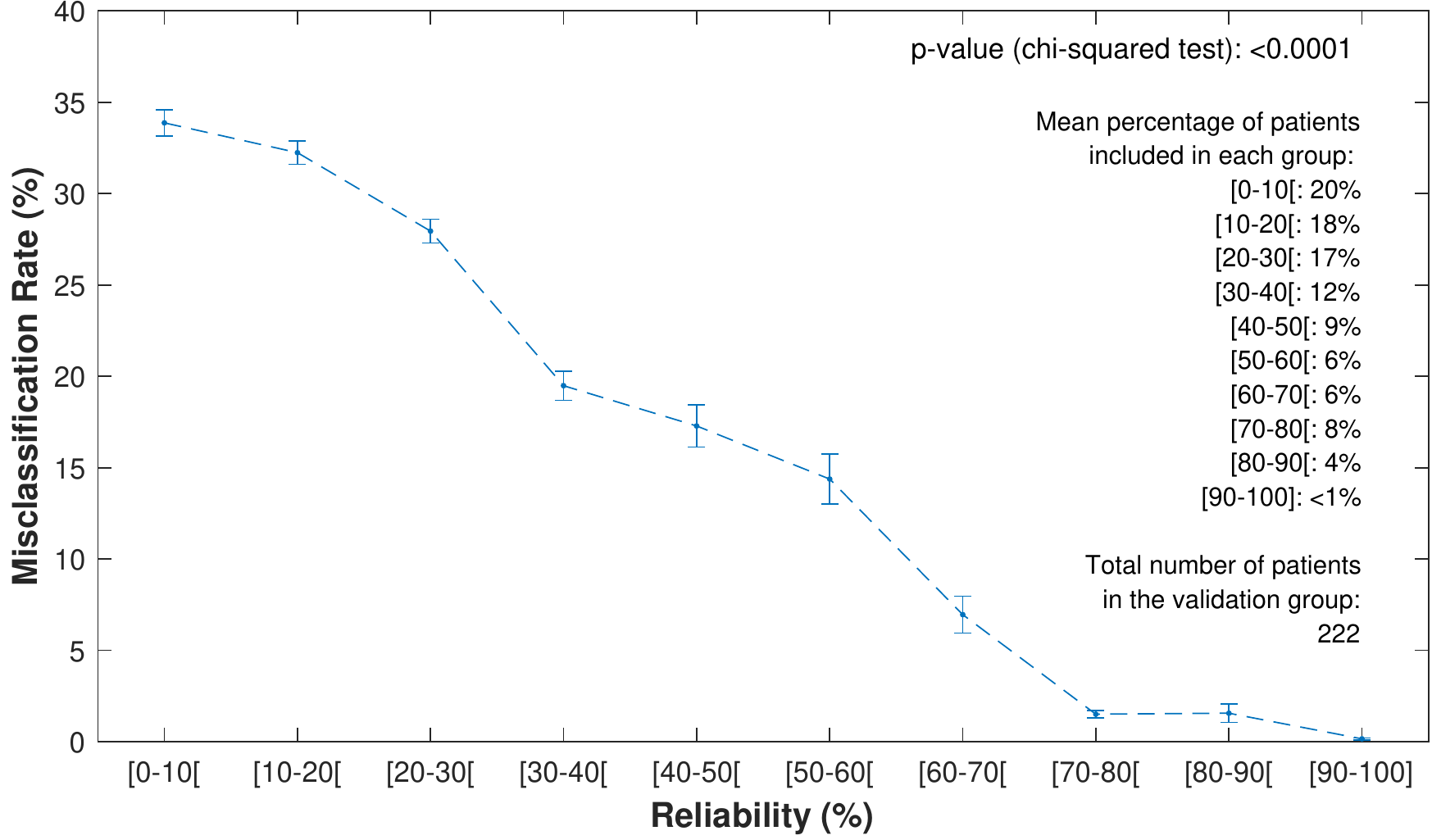}
  \captionof{figure}{Variation of misclassifications rate according to the predicted reliability estimates, for the validation group. The reliability values were divided into subgroup of 10\% for easier interpretation. The points correspond to the mean values and the error bars to the 95\% confidence interval, both computed from the 1000 runs of Monte-Carlo cross-validation. The chi-squared test assesses whether there is an association between the stratified reliability groups (the 10 groups presented in the figure) and the misclassifications, i.e., it tests if the frequency of misclassifications decreases as the reliability increases. The null hypothesis is that the reliability is not associated with the misclassification rate: p-value$<$0.0001, so the null hypothesis is rejected. All samples with a reliability in the range [90-100] were correctly classified, and thus there is no error bar for this group.}
  \label{fig:reliability}
\end{figure}

A visual representation of the relation between the predicted reliability estimates and the rate of misclassifications after the risk stratification is displayed in figure \ref{fig:reliability}, for the validation group. The proposed solution efficiently estimates the reliability of the individual predictions as there is a noticeable relation between such values and the misclassifications number. As expected, higher reliability is highly correlated with a lower forecasting error. Moreover, the obtained results show that for individual predictions with a reliability estimate of 70\% or higher, there is strong evidence that the physicians may trust the algorithm output almost blindly.

No comparison was made with the three other methods (risk score models and standard machine learning ones), as they do not naturally compute a reliability estimation: it is necessary to assess it with an external/auxiliary procedure. Furthermore, to the best of our knowledge, there are no well-established methods (i.e., strongly validated) to estimate the individual reliability of each new sample.
\par
As aforementioned, the calibration curves allow the assessment of the overall reliability, but not the individual one. More specifically, we can only evaluate the expected rate of misclassifications for each level of prediction risk: a different and less personalized, and thus less meaningful, type of reliability. Even so, about this reliability by the level of prediction, it is possible to observe that each model (figures \ref{fig:proposed_calibration} to \ref{fig:ann_calibration}) is more accurate for some risk levels (when the rate of predicted events approaches the rate of observed events), and less accurate for others. Through the inspection of the calibration plots and their characteristics (slope and intercept), it is possible to conclude that the proposed approach seems to present the better overall reliability by prediction risk level: the curve approaches more the ideal one in general, which is supported by a slope closer to 1.

\subsection{Rules interpretability}

As mentioned before, taking the mean value between the virtual patients that represent the negative and positive classes for each feature (the obtained centroids), it is possible to get the thresholds used to create each rule. Therefore, the clinical interpretability of such decision rules is straightforward. Table \ref{tab:threholds} presents the thresholds obtained in our study and the consequent rules.
\par
The obtained rules highly agree with the scoring systems of the ACS risk models previously mentioned (GRACE, TIMI, PURSUIT and GUSTO) about how each feature relates to mortality. This observation supports the idea of clinical interpretability and meaning in the proposed methodology, which facilitates its approval by the physicians.

\begin{table}[H]
\centering
\caption{Outputs advised by the created rules, considering the thresholds obtained from the mean value between the virtual patients. For the continuous variables, the thresholds correspond to the mean value of the 1000 Monte-Carlo runs.}
\label{tab:threholds}
\begin{tabular}{l  c c}
\hline
 & \multicolumn{2}{c}{\textbf{Rules}} \\ \hline
\textbf{Diagnosis} & UA/NSTEMI: survival & STEMI: death \\ 
\textbf{Age} & \textless{} 67 years: survival & $\geq$ 67 years: death \\ 
\textbf{SBP} & $<$ 135 mmHg: death & $\geq$ 135 mmHg: survival \\ 
\textbf{Heart rate} & $<$ 83 bpm: survival & $\geq$ 83 bpm: death \\ 
\textbf{Killip class} & I: survival & II to IV: death \\ 
\textbf{Previous stroke/TIA} & No: survival & Yes: death \\  \hline
\multicolumn{3}{p{\dimexpr\linewidth-2\tabcolsep-2\arrayrulewidth}}{\scriptsize{Note: the levels of the categorical variables (diagnosis, Killip class and previous stroke/TIA) have an ordinal nature from a milder to a more severe condition. Therefore, such variables can be considered as ordinal ones (more specifically, a binary one for the previous stroke/TIA) and the centroids are obtained the same way as for the continuous variables, and consequently the corresponding categorical threshold.}}
\end{tabular}
\end{table}

\section{Discussion}

\subsection{Methodology and results}

Strong interpretability of the clinical knowledge integrated into the proposed approach was one of the main goals in this work. It is easily achievable by the use of dichotomic rules, created based on virtual reference patients (one centroid representing the patients who die and another the patients who survive), and a mean threshold between them. Those binary rules allow to readily identify which variables (observed characteristics) suggest that a given patient may be at high risk. This immediate assessment of critical risk factors for a given patient through binarization is, to the best of your knowledge, a novelty in the risk prediction models.
\par
Then, the interpretability is enhanced by the computation of the acceptance degree expected for each one of those rules, based on the information of previous patients as well. It also gives a sense of personalized medicine to the procedure. The score models present a scoring system that is applied equally to all patients, whereas the machine learning models give different weights to different variables for each patient, such as our approach. However, often that weighting is not easily available/interpretable in ML methodologies, while our model can efficiently show how much each rule is likely to be correct for a given patient.
\par
Therefore, we believe that the proposed approach has an extent of interpretability that is comparable, or even superior, to the risk score models, and far ahead of the standard machine learning models (including the logistic regression one). 
\par
Moreover, the proposed procedure adds the ability to predict the reliability of each particular prediction, which is rarely available to the clinical staff. This estimation can be computed using the same algorithm applied to determine the mortality risk, whereas the approach developed to assess the reliability of already existing score models such as the GRACE one \cite{Myers2020} required an additional method.
\par
As already mentioned, the general reliability of predictions with a given risk can be visually interpreted through calibration curves, when available. Other approaches have been proposed to compute the uncertainty in clinical risk predictions as well \cite{Schetinin2018}. However, such uncertainty is obtained comparing the predicted versus the obtained rate of events and its associated confidence intervals. Therefore, it is estimated a general trustfulness measure of all predictions at the same risk; for example, all the predictions with a risk of 25\% will have the same uncertainty. Contrariwise, the computation of reliability in our methodology is individualized for each prediction, i.e., some predictions with a risk of 25\% can have a low uncertainty and others a high one, depending on the input values. This reliability is then adjustable and more suitable to each prediction.
\par
The obtained results showed that there is a high potentiality to properly evaluate such reliability, estimating when the physicians can deeply trust the algorithm or when more uncertainty in the computed mortality risk is expected. We consider this a very powerful tool to assist the physicians in their decision-making process and to definitely increase their approval for this type of methodology.
\par
Besides accounting for the mentioned novelties, the proposed approach showed the potential to perform well predicting the mortality risk of the patients after ACS and stratifying them into low or high-risk groups. It seems to be able to achieve the good performance of ML methodologies while maintaining higher interpretability. Furthermore, it seems to estimate efficiently the reliability of the individual predictions as well.
\par
Finally, comparing with the standard artificial neural network, the proposed approach obtained better discrimination and calibration. It seems to suggest that the combination of the rule-based methodology with the ANN prediction of rules acceptance requires fewer training samples to reach a given performance and has a better generalization ability than the pure ANN.
\par
Using information retrieved from two different cohorts, only the variables that were present in both datasets were used. Thus, it is expected that the variables being considered are available in the current clinical practice, at least in the Portuguese one. The model generated by the proposed procedure has the advantage to consider a small number of variables - only six were used - which can be obtained in the first 24 hours from admission, allowing to timely perform the advisable treatment. Such variables are already widely considered in existing clinical risk models.
\par
While the score models used in the clinical practice take a fixed number of variables as input, we used a very simple and effective way to impute values when some parameters are missing, which can complement the proposed approach. 
\par
Table \ref{tab:case_study} shows the application of the methodology to real data of a patient admitted with acute coronary syndrome: how the values observed and registered by the clinical staff are converted into the output (considering the threshold) and acceptance (considering the training data) of each rule. For this particular case, five rules (heart rate, diagnosis, previous stroke/TIA, age and, with a lower weight, Killip class) are the ones expected to be more suitable to forecast the patient's condition - personalization; the remaining rule has low predictive power. Then, the outcomes of the conversion of such information into the predicted evaluation are also presented: mortality risk, reliability estimate and risk stratification. The categorization was based on a threshold value of 7\%, which was found to be the one in the training dataset analysis that best splits the patients into low and high-risk groups.

\begin{table}[H]
\centering
\caption{Application of the proposed approach to a real patient admitted with ACS. The rule has an output of 1 if it suggests the patient's death or a value of 0 if it suggests the patient's survival.}
\label{tab:case_study}
\begin{tabular}{l c c c}
\hline
\textbf{\begin{tabular}[c]{@{}c@{}}Feature / Rule\end{tabular}} & \multicolumn{1}{l}{\textbf{\begin{tabular}[c]{@{}c@{}}Observed\\ Value\end{tabular}}} & \multicolumn{1}{l}{\textbf{\begin{tabular}[c]{@{}c@{}}Rule\\ Output\end{tabular}}} & \multicolumn{1}{l}{\textbf{\begin{tabular}[c]{@{}c@{}}Predicted Rule\\ Acceptance\end{tabular}}} \\ \hline
Diagnosis & STEMI & 1 & 87 \% \\ 
Age (years) & 72 & 1 & 75 \% \\ 
SBP (mmHg) & 155 & 0 & 11 \% \\ 
Heart rate (bpm) & 99 & 1 & 91 \% \\ 
Killip class & I & 0 & 59 \% \\ 
Previous stroke/TIA & Yes & 1 & 80 \% \\ \hline
\multicolumn{4}{l}{\textbf{\begin{tabular}[c]{@{}l@{}}Predicted mortality risk: 39\%\\ Predicted reliability: 72\%\\ Stratification: high-risk group (mortality risk $\geq$ 7\%)\end{tabular}}} \\ \hline
\end{tabular}
\end{table}

The threshold optimized by the geometric mean (7\%) approaches the death rate of the studied population (4.95\%). It makes sense that patients with a risk of death higher than a value close to the death rate are considered high-risk. While AUC evaluates the ability to rank the patients according to their absolute risk, the GM assesses the capability to categorize the patients into one of two groups: low or high-risk. The above evidence shows that the threshold used to perform such stratification may depend on the population where it is applied.
\par
Lastly, as the predicted mortality risk is used for such categorization, we can infer that small refinements of such absolute risk may be especially important for the patients whose risk is closer to the threshold of stratification. In fact, it may contribute to define if those patients are included in the low or the high-risk group, and consecutively a different decision (for example, a different treatment) may be applied depending on it.

\subsection{Limitations and future work}

In this study, we introduced a new methodology to incorporate high interpretability in good-performance risk prediction. We do not intend to affirm that our procedure is better or more suitable than others, but rather to present a new method that we believe to have the capability to develop new informative ways to aid the forecasting of patients conditions. Besides, even a Monte-Carlo cross-validation was executed to improve the robustness of the results, it lacks some external validation. In fact, the existing ACS risk score models are well-validated in large cohorts worldwide, while our results are based on a small Portuguese sample size (N=1111). We believe that the application of the proposed approach in such large datasets is imperative not only to validate it but also to improve the overall process, from the selection of features and creation of rules to the prediction of the mortality risk and its associated reliability, due to the larger and more heterogeneous patients information. 
\par
Apart from that, our research presents some limitations that are inherent to almost all the studies in this area. For example, the end-point is the all-cause mortality which is related to any type of death of the patients enrolled in the analysis, whereas the mortality related only to cardiac events would be more suitable, as some of the deaths may not be associated with cardiac issues. Moreover, we only considered the measurements available in the first 24 hours from admission in order to examine the pre-therapy conditions, when it is likely that the combination of some treatments, such as reperfusion and medication (intra-hospital and after discharge), may have had an important impact.
\par
We presented the basis of a new approach that can be further improved. In this work, we selected the risk factors (variables) and then we applied all of them to be used both as rules (the new data labels) and as features to train all the rules acceptance predictors. However, a posterior optimization can be performed, where for example each rule is trained considering only some features and not all of them. This can be seen as an optimization of the multivariate analysis, which is typically performed in score and machine learning models (e.g., using a stepwise selection), maximizing the performance of the method. Moreover, even if the application of the procedure is only presented for the 30-days all-cause mortality, the same approach performed similarly for shorter (14-days) and longer (6-months and 1-year) periods.
\par
Finally, this approach can be applied to many other problems where it may be an appealing and beneficial tool, so it would be interesting to analyze how it performs beyond the ACS scenario, in order to further validate the proposed procedure.

\section{Conclusion}

Clinical decision-making is one of the most important scientific concerns as a misleading evaluation may result in a critical outcome for human health, death in several situations. In this study, we proposed a new decision support system that combines and enhances some of the best properties of both risk score models and machine learning ones into a single method to predict the occurrence risk of a clinical event, and that can be easily implemented in a handheld device or computer. Some of the features incorporated in the described approach allow not only to achieve good performance but also to offer very high interpretability of that system. Additionally, the same procedure is able to compute an individual reliability estimate of such predictions, which is rarely available. The combination of those characteristics (interpretability and reliability) may be peremptory for the clinical acceptance of such a tool, as a wide application of more complex data-driven methodologies is often declined as their "black box" nature minimizes the confidence of the physicians in the generated models.
\par
In this study, the proposed approach was analyzed in the prediction of the mortality risk after acute coronary syndromes, which has great medical importance due to the high cardiovascular death rate worldwide. Furthermore, it is a generic methodology so we believe the same proposed procedure may be successfully applied in other similar scenarios, mainly clinical ones, where the properties described in this study may be of great interest. Even if it requires further validation, it seems to have a significant potential to improve the decision-making process, and not only in ACS events.

\section*{Acknowledgements}

This work was supported by the lookAfterRisk research project (POCI-01-0145-FEDER-030290). The authors would also like to thank the Portuguese Society of Cardiology and the Santa Cruz Hospital (Lisbon, Portugal) for their collaboration, providing the clinical datasets used in this study.

\section*{Conflict of interest statement}
The authors declare no conflicts of interest.

\bibliography{references_paper}

\begin{thebibliography}{51}
\expandafter\ifx\csname natexlab\endcsname\relax\def\natexlab#1{#1}\fi
\providecommand{\url}[1]{\texttt{#1}}
\providecommand{\href}[2]{#2}
\providecommand{\path}[1]{#1}
\providecommand{\DOIprefix}{doi:}
\providecommand{\ArXivprefix}{arXiv:}
\providecommand{\URLprefix}{URL: }
\providecommand{\Pubmedprefix}{pmid:}
\providecommand{\doi}[1]{\href{http://dx.doi.org/#1}{\path{#1}}}
\providecommand{\Pubmed}[1]{\href{pmid:#1}{\path{#1}}}
\providecommand{\bibinfo}[2]{#2}
\ifx\xfnm\relax \def\xfnm[#1]{\unskip,\space#1}\fi
\bibitem[{Sullivan et~al.(2004)Sullivan, Massaro, and
  D'Agostino}]{Sullivan2004}
\bibinfo{author}{L.~M. Sullivan}, \bibinfo{author}{J.~M. Massaro},
  \bibinfo{author}{R.~B. D'Agostino},
\newblock \bibinfo{title}{{Presentation of multivariate data for clinical use:
  The Framingham Study risk score functions}},
\newblock \bibinfo{journal}{Statistics in Medicine} \bibinfo{volume}{23}
  (\bibinfo{year}{2004}) \bibinfo{pages}{1631--1660}.
  \DOIprefix\doi{10.1002/sim.1742}.
\bibitem[{Tay et~al.(2015)Tay, Poh, and Kitney}]{Tay2015}
\bibinfo{author}{D.~Tay}, \bibinfo{author}{C.~L. Poh}, \bibinfo{author}{R.~I.
  Kitney},
\newblock \bibinfo{title}{{A novel neural-inspired learning algorithm with
  application to clinical risk prediction}},
\newblock \bibinfo{journal}{Journal of Biomedical Informatics}
  \bibinfo{volume}{54} (\bibinfo{year}{2015}) \bibinfo{pages}{305--314}.
  \DOIprefix\doi{10.1016/j.jbi.2014.12.014}.
\bibitem[{Chen et~al.(2021)Chen, Shang, Su, Keravnou-Papailiou, Zhao, Antoniou,
  and Shen}]{Chen2020a}
\bibinfo{author}{T.~Chen}, \bibinfo{author}{C.~Shang}, \bibinfo{author}{P.~Su},
  \bibinfo{author}{E.~Keravnou-Papailiou}, \bibinfo{author}{Y.~Zhao},
  \bibinfo{author}{G.~Antoniou}, \bibinfo{author}{Q.~Shen},
\newblock \bibinfo{title}{{A Decision Tree-Initialised Neuro-fuzzy Approach for
  Clinical Decision Support}},
\newblock \bibinfo{journal}{Artificial Intelligence in Medicine}
  \bibinfo{volume}{111} (\bibinfo{year}{2021}) \bibinfo{pages}{101986}.
  \DOIprefix\doi{10.1016/j.artmed.2020.101986}.
\bibitem[{Davoodi and Moradi(2018)}]{Davoodi2018}
\bibinfo{author}{R.~Davoodi}, \bibinfo{author}{M.~H. Moradi},
\newblock \bibinfo{title}{{Mortality prediction in intensive care units (ICUs)
  using a deep rule-based fuzzy classifier}},
\newblock \bibinfo{journal}{Journal of Biomedical Informatics}
  \bibinfo{volume}{79} (\bibinfo{year}{2018}) \bibinfo{pages}{48--59}.
  \DOIprefix\doi{10.1016/j.jbi.2018.02.008}.
\bibitem[{Obermeyer and Emanuel(2016)}]{Obermeyer2016}
\bibinfo{author}{Z.~Obermeyer}, \bibinfo{author}{E.~J. Emanuel},
\newblock \bibinfo{title}{{Predicting the Future — Big Data, Machine
  Learning, and Clinical Medicine}},
\newblock \bibinfo{journal}{New England Journal of Medicine}
  \bibinfo{volume}{375} (\bibinfo{year}{2016}) \bibinfo{pages}{1216--1219}.
  \DOIprefix\doi{10.1056/NEJMp1606181}.
\bibitem[{Watson et~al.(2019)Watson, Krutzinna, Bruce, Griffiths, McInnes,
  Barnes, and Floridi}]{Watson2019}
\bibinfo{author}{D.~S. Watson}, \bibinfo{author}{J.~Krutzinna},
  \bibinfo{author}{I.~N. Bruce}, \bibinfo{author}{C.~E. Griffiths},
  \bibinfo{author}{I.~B. McInnes}, \bibinfo{author}{M.~R. Barnes},
  \bibinfo{author}{L.~Floridi},
\newblock \bibinfo{title}{{Clinical applications of machine learning
  algorithms: beyond the black box}},
\newblock \bibinfo{journal}{BMJ} \bibinfo{volume}{364} (\bibinfo{year}{2019})
  \bibinfo{pages}{l886}. \DOIprefix\doi{10.1136/bmj.l886}.
\bibitem[{Murdoch et~al.(2019)Murdoch, Singh, Kumbier, Abbasi-Asl, and
  Yu}]{Murdoch2019}
\bibinfo{author}{W.~J. Murdoch}, \bibinfo{author}{C.~Singh},
  \bibinfo{author}{K.~Kumbier}, \bibinfo{author}{R.~Abbasi-Asl},
  \bibinfo{author}{B.~Yu},
\newblock \bibinfo{title}{{Definitions, methods, and applications in
  interpretable machine learning}},
\newblock \bibinfo{journal}{Proceedings of the National Academy of Sciences}
  \bibinfo{volume}{116} (\bibinfo{year}{2019}) \bibinfo{pages}{22071--22080}.
  \DOIprefix\doi{10.1073/pnas.1900654116}.
\bibitem[{Miller(2019)}]{Miller2019}
\bibinfo{author}{T.~Miller},
\newblock \bibinfo{title}{{Explanation in artificial intelligence: Insights
  from the social sciences}},
\newblock \bibinfo{journal}{Artificial Intelligence} \bibinfo{volume}{267}
  (\bibinfo{year}{2019}) \bibinfo{pages}{1--38}.
  \DOIprefix\doi{10.1016/j.artint.2018.07.007}.
\bibitem[{Guidotti et~al.(2019)Guidotti, Monreale, Ruggieri, Turini, Giannotti,
  and Pedreschi}]{10.1145/3236009}
\bibinfo{author}{R.~Guidotti}, \bibinfo{author}{A.~Monreale},
  \bibinfo{author}{S.~Ruggieri}, \bibinfo{author}{F.~Turini},
  \bibinfo{author}{F.~Giannotti}, \bibinfo{author}{D.~Pedreschi},
\newblock \bibinfo{title}{{A Survey of Methods for Explaining Black Box
  Models}},
\newblock \bibinfo{journal}{ACM Computing Surveys} \bibinfo{volume}{51}
  (\bibinfo{year}{2019}) \bibinfo{pages}{1--42}.
  \DOIprefix\doi{10.1145/3236009}.
\bibitem[{Allen et~al.(2017)Allen, Matlock, Shetterly, Xu, Levy, Portalupi,
  McIlvennan, Gurwitz, Johnson, Smith, and Magid}]{Allen2017}
\bibinfo{author}{L.~A. Allen}, \bibinfo{author}{D.~D. Matlock},
  \bibinfo{author}{S.~M. Shetterly}, \bibinfo{author}{S.~Xu},
  \bibinfo{author}{W.~C. Levy}, \bibinfo{author}{L.~B. Portalupi},
  \bibinfo{author}{C.~K. McIlvennan}, \bibinfo{author}{J.~H. Gurwitz},
  \bibinfo{author}{E.~S. Johnson}, \bibinfo{author}{D.~H. Smith},
  \bibinfo{author}{D.~J. Magid},
\newblock \bibinfo{title}{{Use of Risk Models to Predict Death in the Next Year
  Among Individual Ambulatory Patients With Heart Failure}},
\newblock \bibinfo{journal}{JAMA Cardiology} \bibinfo{volume}{2}
  (\bibinfo{year}{2017}) \bibinfo{pages}{435}.
  \DOIprefix\doi{10.1001/jamacardio.2016.5036}.
\bibitem[{Li et~al.(2019)Li, Sperrin, Belmonte, Pate, Ashcroft, and van
  Staa}]{Li2019}
\bibinfo{author}{Y.~Li}, \bibinfo{author}{M.~Sperrin},
  \bibinfo{author}{M.~Belmonte}, \bibinfo{author}{A.~Pate},
  \bibinfo{author}{D.~M. Ashcroft}, \bibinfo{author}{T.~P. van Staa},
\newblock \bibinfo{title}{{Do population-level risk prediction models that use
  routinely collected health data reliably predict individual risks?}},
\newblock \bibinfo{journal}{Scientific Reports} \bibinfo{volume}{9}
  (\bibinfo{year}{2019}) \bibinfo{pages}{11222}.
  \DOIprefix\doi{10.1038/s41598-019-47712-5}.
\bibitem[{Bosni{\'{c}} and Kononenko(2008)}]{Bosnic2008a}
\bibinfo{author}{Z.~Bosni{\'{c}}}, \bibinfo{author}{I.~Kononenko},
\newblock \bibinfo{title}{{Comparison of approaches for estimating reliability
  of individual regression predictions}},
\newblock \bibinfo{journal}{Data {\&} Knowledge Engineering}
  \bibinfo{volume}{67} (\bibinfo{year}{2008}) \bibinfo{pages}{504--516}.
  \DOIprefix\doi{10.1016/j.datak.2008.08.001}.
\bibitem[{Jiang et~al.(2018)Jiang, Kim, Guan, and Gupta}]{Jiang2018}
\bibinfo{author}{H.~Jiang}, \bibinfo{author}{B.~Kim}, \bibinfo{author}{M.~Y.
  Guan}, \bibinfo{author}{M.~Gupta},
\newblock \bibinfo{title}{{To Trust or Not to Trust a Classifier}},
\newblock in: \bibinfo{booktitle}{Proceedings of the 32nd International
  Conference on Neural Information Processing Systems}, NIPS'18,
  \bibinfo{publisher}{Curran Associates Inc.}, \bibinfo{address}{Red Hook, NY,
  USA}, \bibinfo{year}{2018}, pp. \bibinfo{pages}{5546--5557}.
\bibitem[{Schulam and Saria(2019)}]{Schulam2019}
\bibinfo{author}{P.~Schulam}, \bibinfo{author}{S.~Saria},
\newblock \bibinfo{title}{{Can You Trust This Prediction? Auditing Pointwise
  Reliability After Learning}},
\newblock in: \bibinfo{editor}{K.~Chaudhuri}, \bibinfo{editor}{M.~Sugiyama}
  (Eds.), \bibinfo{booktitle}{Proceedings of Machine Learning Research},
  volume~\bibinfo{volume}{89} of \textit{\bibinfo{series}{Proceedings of
  Machine Learning Research}}, \bibinfo{publisher}{PMLR}, \bibinfo{year}{2019},
  pp. \bibinfo{pages}{1022--1031}.
\bibitem[{Kailkhura et~al.(2019)Kailkhura, Gallagher, Kim, Hiszpanski, and
  Han}]{Kailkhura2019}
\bibinfo{author}{B.~Kailkhura}, \bibinfo{author}{B.~Gallagher},
  \bibinfo{author}{S.~Kim}, \bibinfo{author}{A.~Hiszpanski},
  \bibinfo{author}{T.~Y.-J. Han},
\newblock \bibinfo{title}{{Reliable and explainable machine-learning methods
  for accelerated material discovery}},
\newblock \bibinfo{journal}{npj Computational Materials} \bibinfo{volume}{5}
  (\bibinfo{year}{2019}) \bibinfo{pages}{108}.
  \DOIprefix\doi{10.1038/s41524-019-0248-2}.
\bibitem[{Myers et~al.(2020)Myers, Ng, Severson, Kartoun, Dai, Huang, Anderson,
  and Stultz}]{Myers2020}
\bibinfo{author}{P.~D. Myers}, \bibinfo{author}{K.~Ng},
  \bibinfo{author}{K.~Severson}, \bibinfo{author}{U.~Kartoun},
  \bibinfo{author}{W.~Dai}, \bibinfo{author}{W.~Huang}, \bibinfo{author}{F.~A.
  Anderson}, \bibinfo{author}{C.~M. Stultz},
\newblock \bibinfo{title}{{Identifying unreliable predictions in clinical risk
  models}},
\newblock \bibinfo{journal}{npj Digital Medicine} \bibinfo{volume}{3}
  (\bibinfo{year}{2020}) \bibinfo{pages}{8}.
  \DOIprefix\doi{10.1038/s41746-019-0209-7}.
\bibitem[{{World Health Organization}(2018)}]{WorldHealthOrganization2018}
\bibinfo{author}{{World Health Organization}}, \bibinfo{title}{{The top 10
  causes of death}}, \bibinfo{year}{2018}. \URLprefix
  \url{https://www.who.int/en/news-room/fact-sheets/detail/the-top-10-causes-of-death}.
\bibitem[{Boersma et~al.(2000)Boersma, Pieper, Steyerberg, Wilcox, Chang, Lee,
  Akkerhuis, Harrington, Deckers, Armstrong, Lincoff, Califf, Topol, and
  Simoons}]{Boersma2000}
\bibinfo{author}{E.~Boersma}, \bibinfo{author}{K.~S. Pieper},
  \bibinfo{author}{E.~W. Steyerberg}, \bibinfo{author}{R.~G. Wilcox},
  \bibinfo{author}{W.-C. Chang}, \bibinfo{author}{K.~L. Lee},
  \bibinfo{author}{K.~M. Akkerhuis}, \bibinfo{author}{R.~A. Harrington},
  \bibinfo{author}{J.~W. Deckers}, \bibinfo{author}{P.~W. Armstrong},
  \bibinfo{author}{A.~M. Lincoff}, \bibinfo{author}{R.~M. Califf},
  \bibinfo{author}{E.~J. Topol}, \bibinfo{author}{M.~L. Simoons},
\newblock \bibinfo{title}{{Predictors of Outcome in Patients With Acute
  Coronary Syndromes Without Persistent ST-Segment Elevation}},
\newblock \bibinfo{journal}{Circulation} \bibinfo{volume}{101}
  (\bibinfo{year}{2000}) \bibinfo{pages}{2557--2567}.
  \DOIprefix\doi{10.1161/01.CIR.101.22.2557}.
\bibitem[{Antman et~al.(2000)Antman, Cohen, Bernink, McCabe, Horacek, Papuchis,
  Mautner, Corbalan, Radley, and Braunwald}]{Antman2000a}
\bibinfo{author}{E.~M. Antman}, \bibinfo{author}{M.~Cohen},
  \bibinfo{author}{P.~J. L.~M. Bernink}, \bibinfo{author}{C.~H. McCabe},
  \bibinfo{author}{T.~Horacek}, \bibinfo{author}{G.~Papuchis},
  \bibinfo{author}{B.~Mautner}, \bibinfo{author}{R.~Corbalan},
  \bibinfo{author}{D.~Radley}, \bibinfo{author}{E.~Braunwald},
\newblock \bibinfo{title}{{The TIMI Risk Score for Unstable Angina/Non–ST
  Elevation MI}},
\newblock \bibinfo{journal}{JAMA} \bibinfo{volume}{284} (\bibinfo{year}{2000})
  \bibinfo{pages}{835}. \DOIprefix\doi{10.1001/jama.284.7.835}.
\bibitem[{Morrow et~al.(2000)Morrow, Antman, Charlesworth, Cairns, Murphy,
  de~Lemos, Giugliano, McCabe, and Braunwald}]{Morrow2000a}
\bibinfo{author}{D.~A. Morrow}, \bibinfo{author}{E.~M. Antman},
  \bibinfo{author}{A.~Charlesworth}, \bibinfo{author}{R.~Cairns},
  \bibinfo{author}{S.~A. Murphy}, \bibinfo{author}{J.~A. de~Lemos},
  \bibinfo{author}{R.~P. Giugliano}, \bibinfo{author}{C.~H. McCabe},
  \bibinfo{author}{E.~Braunwald},
\newblock \bibinfo{title}{{TIMI Risk Score for ST-Elevation Myocardial
  Infarction: A Convenient, Bedside, Clinical Score for Risk Assessment at
  Presentation}},
\newblock \bibinfo{journal}{Circulation} \bibinfo{volume}{102}
  (\bibinfo{year}{2000}) \bibinfo{pages}{2031--2037}.
  \DOIprefix\doi{10.1161/01.CIR.102.17.2031}.
\bibitem[{Lee et~al.(1995)Lee, Woodlief, Topol, Weaver, Betriu, Col, Simoons,
  Aylward, {Van de Werf}, and Califf}]{L.1995}
\bibinfo{author}{K.~L. Lee}, \bibinfo{author}{L.~H. Woodlief},
  \bibinfo{author}{E.~J. Topol}, \bibinfo{author}{W.~D. Weaver},
  \bibinfo{author}{A.~Betriu}, \bibinfo{author}{J.~Col},
  \bibinfo{author}{M.~Simoons}, \bibinfo{author}{P.~Aylward},
  \bibinfo{author}{F.~{Van de Werf}}, \bibinfo{author}{R.~M. Califf},
\newblock \bibinfo{title}{{Predictors of 30-Day Mortality in the Era of
  Reperfusion for Acute Myocardial Infarction}},
\newblock \bibinfo{journal}{Circulation} \bibinfo{volume}{91}
  (\bibinfo{year}{1995}) \bibinfo{pages}{1659--1668}.
  \DOIprefix\doi{10.1161/01.CIR.91.6.1659}.
\bibitem[{Armstrong et~al.(1998)Armstrong, Fu, Chang, Topol, Granger, Betriu,
  {Van de Werf}, Lee, and Califf}]{Armstrong1998a}
\bibinfo{author}{P.~W. Armstrong}, \bibinfo{author}{Y.~Fu},
  \bibinfo{author}{W.-C. Chang}, \bibinfo{author}{E.~J. Topol},
  \bibinfo{author}{C.~B. Granger}, \bibinfo{author}{A.~Betriu},
  \bibinfo{author}{F.~{Van de Werf}}, \bibinfo{author}{K.~L. Lee},
  \bibinfo{author}{R.~M. Califf},
\newblock \bibinfo{title}{{Acute Coronary Syndromes in the GUSTO-IIb Trial}},
\newblock \bibinfo{journal}{Circulation} \bibinfo{volume}{98}
  (\bibinfo{year}{1998}) \bibinfo{pages}{1860--1868}.
  \DOIprefix\doi{10.1161/01.CIR.98.18.1860}.
\bibitem[{Granger(2003)}]{Granger2003}
\bibinfo{author}{C.~B. Granger},
\newblock \bibinfo{title}{{Predictors of Hospital Mortality in the Global
  Registry of Acute Coronary Events}},
\newblock \bibinfo{journal}{Archives of Internal Medicine}
  \bibinfo{volume}{163} (\bibinfo{year}{2003}) \bibinfo{pages}{2345}.
  \DOIprefix\doi{10.1001/archinte.163.19.2345}.
\bibitem[{Fox et~al.(2006)Fox, Dabbous, Goldberg, Pieper, Eagle, {Van de Werf},
  Avezum, Goodman, Flather, Anderson, and Granger}]{Fox2006}
\bibinfo{author}{K.~A.~A. Fox}, \bibinfo{author}{O.~H. Dabbous},
  \bibinfo{author}{R.~J. Goldberg}, \bibinfo{author}{K.~S. Pieper},
  \bibinfo{author}{K.~A. Eagle}, \bibinfo{author}{F.~{Van de Werf}},
  \bibinfo{author}{{\'{A}}.~Avezum}, \bibinfo{author}{S.~G. Goodman},
  \bibinfo{author}{M.~D. Flather}, \bibinfo{author}{F.~A. Anderson},
  \bibinfo{author}{C.~B. Granger},
\newblock \bibinfo{title}{{Prediction of risk of death and myocardial
  infarction in the six months after presentation with acute coronary syndrome:
  prospective multinational observational study (GRACE)}},
\newblock \bibinfo{journal}{BMJ} \bibinfo{volume}{333} (\bibinfo{year}{2006})
  \bibinfo{pages}{1091}. \DOIprefix\doi{10.1136/bmj.38985.646481.55}.
\bibitem[{Huang et~al.(2018)Huang, Dong, Duan, and Liu}]{Huang2018}
\bibinfo{author}{Z.~Huang}, \bibinfo{author}{W.~Dong},
  \bibinfo{author}{H.~Duan}, \bibinfo{author}{J.~Liu},
\newblock \bibinfo{title}{{A Regularized Deep Learning Approach for Clinical
  Risk Prediction of Acute Coronary Syndrome Using Electronic Health Records}},
\newblock \bibinfo{journal}{IEEE Transactions on Biomedical Engineering}
  \bibinfo{volume}{65} (\bibinfo{year}{2018}) \bibinfo{pages}{956--968}.
  \DOIprefix\doi{10.1109/TBME.2017.2731158}.
\bibitem[{Myers et~al.(2017)Myers, Scirica, and Stultz}]{Myers2017}
\bibinfo{author}{P.~D. Myers}, \bibinfo{author}{B.~M. Scirica},
  \bibinfo{author}{C.~M. Stultz},
\newblock \bibinfo{title}{{Machine Learning Improves Risk Stratification After
  Acute Coronary Syndrome}},
\newblock \bibinfo{journal}{Scientific Reports} \bibinfo{volume}{7}
  (\bibinfo{year}{2017}) \bibinfo{pages}{12692}.
  \DOIprefix\doi{10.1038/s41598-017-12951-x}.
\bibitem[{Shouval et~al.(2017)Shouval, Hadanny, Shlomo, Iakobishvili, Unger,
  Zahger, Alcalai, Atar, Gottlieb, Matetzky, Goldenberg, and
  Beigel}]{Shouval2017}
\bibinfo{author}{R.~Shouval}, \bibinfo{author}{A.~Hadanny},
  \bibinfo{author}{N.~Shlomo}, \bibinfo{author}{Z.~Iakobishvili},
  \bibinfo{author}{R.~Unger}, \bibinfo{author}{D.~Zahger},
  \bibinfo{author}{R.~Alcalai}, \bibinfo{author}{S.~Atar},
  \bibinfo{author}{S.~Gottlieb}, \bibinfo{author}{S.~Matetzky},
  \bibinfo{author}{I.~Goldenberg}, \bibinfo{author}{R.~Beigel},
\newblock \bibinfo{title}{{Machine learning for prediction of 30-day mortality
  after ST elevation myocardial infraction: An Acute Coronary Syndrome Israeli
  Survey data mining study}},
\newblock \bibinfo{journal}{International Journal of Cardiology}
  \bibinfo{volume}{246} (\bibinfo{year}{2017}) \bibinfo{pages}{7--13}.
  \DOIprefix\doi{10.1016/j.ijcard.2017.05.067}.
\bibitem[{Kwon et~al.(2019)Kwon, Jeon, Kim, Kim, Lim, Kim, Song, Park, Choi,
  and Oh}]{Kwon2019}
\bibinfo{author}{J.-m. Kwon}, \bibinfo{author}{K.-H. Jeon},
  \bibinfo{author}{H.~M. Kim}, \bibinfo{author}{M.~J. Kim},
  \bibinfo{author}{S.~Lim}, \bibinfo{author}{K.-H. Kim}, \bibinfo{author}{P.~S.
  Song}, \bibinfo{author}{J.~Park}, \bibinfo{author}{R.~K. Choi},
  \bibinfo{author}{B.-H. Oh},
\newblock \bibinfo{title}{{Deep-learning-based risk stratification for
  mortality of patients with acute myocardial infarction}},
\newblock \bibinfo{journal}{PLOS ONE} \bibinfo{volume}{14}
  (\bibinfo{year}{2019}) \bibinfo{pages}{e0224502}.
  \DOIprefix\doi{10.1371/journal.pone.0224502}.
\bibitem[{Myers et~al.(2019)Myers, Huang, Anderson, and Stultz}]{Myers2019}
\bibinfo{author}{P.~D. Myers}, \bibinfo{author}{W.~Huang},
  \bibinfo{author}{F.~Anderson}, \bibinfo{author}{C.~M. Stultz},
\newblock \bibinfo{title}{{Choosing Clinical Variables for Risk Stratification
  Post-Acute Coronary Syndrome}},
\newblock \bibinfo{journal}{Scientific Reports} \bibinfo{volume}{9}
  (\bibinfo{year}{2019}) \bibinfo{pages}{14631}.
  \DOIprefix\doi{10.1038/s41598-019-50933-3}.
\bibitem[{Zou and Hastie(2005)}]{Zou2005}
\bibinfo{author}{H.~Zou}, \bibinfo{author}{T.~Hastie},
\newblock \bibinfo{title}{{Regularization and Variable Selection via the
  Elastic Net}},
\newblock \bibinfo{journal}{Journal of the Royal Statistical Society. Series B
  (Statistical Methodology)} \bibinfo{volume}{67} (\bibinfo{year}{2005})
  \bibinfo{pages}{301--320}.
\bibitem[{Burden and Winkler(2008)}]{Burden2008}
\bibinfo{author}{F.~Burden}, \bibinfo{author}{D.~Winkler},
\newblock \bibinfo{title}{{Bayesian Regularization of Neural Networks}},
\newblock in: \bibinfo{booktitle}{Methods in molecular biology (Clifton,
  N.J.)}, volume \bibinfo{volume}{458}, \bibinfo{year}{2008}, pp.
  \bibinfo{pages}{23--42}. \DOIprefix\doi{10.1007/978-1-60327-101-1_3}.
\bibitem[{{Van Calster} et~al.(2019){Van Calster}, McLernon, van Smeden,
  Wynants, and Steyerberg}]{VanCalster2019}
\bibinfo{author}{B.~{Van Calster}}, \bibinfo{author}{D.~J. McLernon},
  \bibinfo{author}{M.~van Smeden}, \bibinfo{author}{L.~Wynants},
  \bibinfo{author}{E.~W. Steyerberg},
\newblock \bibinfo{title}{{Calibration: the Achilles heel of predictive
  analytics}},
\newblock \bibinfo{journal}{BMC Medicine} \bibinfo{volume}{17}
  (\bibinfo{year}{2019}) \bibinfo{pages}{230}.
  \DOIprefix\doi{10.1186/s12916-019-1466-7}.
\bibitem[{Wynants et~al.(2019)Wynants, van Smeden, McLernon, Timmerman,
  Steyerberg, and {Van Calster}}]{Wynants2019}
\bibinfo{author}{L.~Wynants}, \bibinfo{author}{M.~van Smeden},
  \bibinfo{author}{D.~J. McLernon}, \bibinfo{author}{D.~Timmerman},
  \bibinfo{author}{E.~W. Steyerberg}, \bibinfo{author}{B.~{Van Calster}},
\newblock \bibinfo{title}{{Three myths about risk thresholds for prediction
  models}},
\newblock \bibinfo{journal}{BMC Medicine} \bibinfo{volume}{17}
  (\bibinfo{year}{2019}) \bibinfo{pages}{192}.
  \DOIprefix\doi{10.1186/s12916-019-1425-3}.
\bibitem[{Zou et~al.(2016)Zou, Xie, Lin, Wu, and Ju}]{Zou2016}
\bibinfo{author}{Q.~Zou}, \bibinfo{author}{S.~Xie}, \bibinfo{author}{Z.~Lin},
  \bibinfo{author}{M.~Wu}, \bibinfo{author}{Y.~Ju},
\newblock \bibinfo{title}{{Finding the Best Classification Threshold in
  Imbalanced Classification}},
\newblock \bibinfo{journal}{Big Data Research} \bibinfo{volume}{5}
  (\bibinfo{year}{2016}) \bibinfo{pages}{2--8}.
  \DOIprefix\doi{10.1016/j.bdr.2015.12.001}.
\bibitem[{Wasserstein and Lazar(2016)}]{Wasserstein2016}
\bibinfo{author}{R.~L. Wasserstein}, \bibinfo{author}{N.~A. Lazar},
\newblock \bibinfo{title}{{The ASA Statement on p -Values: Context, Process,
  and Purpose}},
\newblock \bibinfo{journal}{The American Statistician} \bibinfo{volume}{70}
  (\bibinfo{year}{2016}) \bibinfo{pages}{129--133}.
  \DOIprefix\doi{10.1080/00031305.2016.1154108}.
\bibitem[{Gavish et~al.(2008)Gavish, Ben-Dov, and Bursztyn}]{Gavish2008}
\bibinfo{author}{B.~Gavish}, \bibinfo{author}{I.~Z. Ben-Dov},
  \bibinfo{author}{M.~Bursztyn},
\newblock \bibinfo{title}{{Linear relationship between systolic and diastolic
  blood pressure monitored over 24 h: assessment and correlates}},
\newblock \bibinfo{journal}{Journal of Hypertension} \bibinfo{volume}{26}
  (\bibinfo{year}{2008}) \bibinfo{pages}{199--209}.
  \DOIprefix\doi{10.1097/HJH.0b013e3282f25b5a}.
\bibitem[{{de Ara{\'{u}}jo Gon{\c{c}}alves} et~al.(2005){de Ara{\'{u}}jo
  Gon{\c{c}}alves}, Ferreira, Aguiar, and Seabra-Gomes}]{DeAraujoGoncalves2005}
\bibinfo{author}{P.~{de Ara{\'{u}}jo Gon{\c{c}}alves}},
  \bibinfo{author}{J.~Ferreira}, \bibinfo{author}{C.~Aguiar},
  \bibinfo{author}{R.~Seabra-Gomes},
\newblock \bibinfo{title}{{TIMI, PURSUIT, and GRACE risk scores: sustained
  prognostic value and interaction with revascularization in NSTE‐ACS}},
\newblock \bibinfo{journal}{European Heart Journal} \bibinfo{volume}{26}
  (\bibinfo{year}{2005}) \bibinfo{pages}{865--872}.
  \DOIprefix\doi{10.1093/eurheartj/ehi187}.
\bibitem[{Billiet et~al.(2018)Billiet, {Van Huffel}, and {Van
  Belle}}]{Billiet2018}
\bibinfo{author}{L.~Billiet}, \bibinfo{author}{S.~{Van Huffel}},
  \bibinfo{author}{V.~{Van Belle}},
\newblock \bibinfo{title}{{Interval Coded Scoring: a toolbox for interpretable
  scoring systems}},
\newblock \bibinfo{journal}{PeerJ Computer Science} \bibinfo{volume}{4}
  (\bibinfo{year}{2018}) \bibinfo{pages}{e150}.
  \DOIprefix\doi{10.7717/peerj-cs.150}.
\bibitem[{Caicedo-Torres and Gutierrez(2019)}]{Caicedo-Torres2019}
\bibinfo{author}{W.~Caicedo-Torres}, \bibinfo{author}{J.~Gutierrez},
\newblock \bibinfo{title}{{ISeeU: Visually interpretable deep learning for
  mortality prediction inside the ICU}},
\newblock \bibinfo{journal}{Journal of Biomedical Informatics}
  \bibinfo{volume}{98} (\bibinfo{year}{2019}) \bibinfo{pages}{103269}.
  \DOIprefix\doi{https://doi.org/10.1016/j.jbi.2019.103269}.
\bibitem[{Chen et~al.(2020)Chen, Dong, Wang, Lu, Kaymak, and Huang}]{Chen2020}
\bibinfo{author}{P.~Chen}, \bibinfo{author}{W.~Dong},
  \bibinfo{author}{J.~Wang}, \bibinfo{author}{X.~Lu},
  \bibinfo{author}{U.~Kaymak}, \bibinfo{author}{Z.~Huang},
\newblock \bibinfo{title}{{Interpretable clinical prediction via
  attention-based neural network}},
\newblock \bibinfo{journal}{BMC Medical Informatics and Decision Making}
  \bibinfo{volume}{20} (\bibinfo{year}{2020}) \bibinfo{pages}{131}.
  \DOIprefix\doi{10.1186/s12911-020-1110-7}.
\bibitem[{Caruana et~al.(2015)Caruana, Lou, Gehrke, Koch, Sturm, and
  Elhadad}]{Caruana2015}
\bibinfo{author}{R.~Caruana}, \bibinfo{author}{Y.~Lou},
  \bibinfo{author}{J.~Gehrke}, \bibinfo{author}{P.~Koch},
  \bibinfo{author}{M.~Sturm}, \bibinfo{author}{N.~Elhadad},
\newblock \bibinfo{title}{{Intelligible Models for HealthCare: Predicting
  Pneumonia Risk and Hospital 30-Day Readmission}},
\newblock in: \bibinfo{booktitle}{Proceedings of the 21th ACM SIGKDD
  International Conference on Knowledge Discovery and Data Mining}, KDD '15,
  \bibinfo{publisher}{Association for Computing Machinery},
  \bibinfo{address}{New York, NY, USA}, \bibinfo{year}{2015}, pp.
  \bibinfo{pages}{1721--1730}. \DOIprefix\doi{10.1145/2783258.2788613}.
\bibitem[{Gu et~al.(2020)Gu, Su, and Zhao}]{Gu2020}
\bibinfo{author}{D.~Gu}, \bibinfo{author}{K.~Su}, \bibinfo{author}{H.~Zhao},
\newblock \bibinfo{title}{{A case-based ensemble learning system for
  explainable breast cancer recurrence prediction}},
\newblock \bibinfo{journal}{Artificial Intelligence in Medicine}
  \bibinfo{volume}{107} (\bibinfo{year}{2020}) \bibinfo{pages}{101858}.
  \DOIprefix\doi{10.1016/j.artmed.2020.101858}.
\bibitem[{Fernandes et~al.(2020)Fernandes, Vieira, Leite, Palos, Finkelstein,
  and Sousa}]{Fernandes2020}
\bibinfo{author}{M.~Fernandes}, \bibinfo{author}{S.~M. Vieira},
  \bibinfo{author}{F.~Leite}, \bibinfo{author}{C.~Palos},
  \bibinfo{author}{S.~Finkelstein}, \bibinfo{author}{J.~M. Sousa},
\newblock \bibinfo{title}{{Clinical Decision Support Systems for Triage in the
  Emergency Department using Intelligent Systems: a Review}},
\newblock \bibinfo{journal}{Artificial Intelligence in Medicine}
  \bibinfo{volume}{102} (\bibinfo{year}{2020}) \bibinfo{pages}{101762}.
  \DOIprefix\doi{10.1016/j.artmed.2019.101762}.
\bibitem[{Artetxe et~al.(2018)Artetxe, Beristain, and
  Gra{\~{n}}a}]{Artetxe2018}
\bibinfo{author}{A.~Artetxe}, \bibinfo{author}{A.~Beristain},
  \bibinfo{author}{M.~Gra{\~{n}}a},
\newblock \bibinfo{title}{{Predictive models for hospital readmission risk: A
  systematic review of methods}},
\newblock \bibinfo{journal}{Computer Methods and Programs in Biomedicine}
  \bibinfo{volume}{164} (\bibinfo{year}{2018}) \bibinfo{pages}{49--64}.
  \DOIprefix\doi{10.1016/j.cmpb.2018.06.006}.
\bibitem[{Barra et~al.(2012)Barra, Provid{\^{e}}ncia, Paiva, Caetano, Almeida,
  Gomes, and Marques}]{Barra2012}
\bibinfo{author}{S.~Barra}, \bibinfo{author}{R.~Provid{\^{e}}ncia},
  \bibinfo{author}{L.~Paiva}, \bibinfo{author}{F.~Caetano},
  \bibinfo{author}{I.~Almeida}, \bibinfo{author}{P.~Gomes},
  \bibinfo{author}{A.~L. Marques},
\newblock \bibinfo{title}{{ACHTUNG-Rule: A new and improved model for
  prognostic assessment in myocardial infarction}},
\newblock \bibinfo{journal}{European Heart Journal: Acute Cardiovascular Care}
  \bibinfo{volume}{1} (\bibinfo{year}{2012}) \bibinfo{pages}{320--336}.
  \DOIprefix\doi{10.1177/2048872612466536}.
\bibitem[{Elbarouni et~al.(2009)Elbarouni, Goodman, Yan, Welsh, Kornder,
  DeYoung, Wong, Rose, Grondin, Gallo, Tan, Casanova, Eagle, and
  Yan}]{Elbarouni2009}
\bibinfo{author}{B.~Elbarouni}, \bibinfo{author}{S.~G. Goodman},
  \bibinfo{author}{R.~T. Yan}, \bibinfo{author}{R.~C. Welsh},
  \bibinfo{author}{J.~M. Kornder}, \bibinfo{author}{J.~P. DeYoung},
  \bibinfo{author}{G.~C. Wong}, \bibinfo{author}{B.~Rose},
  \bibinfo{author}{F.~R. Grondin}, \bibinfo{author}{R.~Gallo},
  \bibinfo{author}{M.~Tan}, \bibinfo{author}{A.~Casanova},
  \bibinfo{author}{K.~A. Eagle}, \bibinfo{author}{A.~T. Yan},
\newblock \bibinfo{title}{{Validation of the Global Registry of Acute Coronary
  Event (GRACE) risk score for in-hospital mortality in patients with acute
  coronary syndrome in Canada}},
\newblock \bibinfo{journal}{American Heart Journal} \bibinfo{volume}{158}
  (\bibinfo{year}{2009}) \bibinfo{pages}{392--399}.
  \DOIprefix\doi{10.1016/j.ahj.2009.06.010}.
\bibitem[{Yan et~al.(2007)Yan, Yan, Tan, Casanova, Labinaz, Sridhar, Fitchett,
  Langer, and Goodman}]{Yan2007}
\bibinfo{author}{A.~T. Yan}, \bibinfo{author}{R.~T. Yan},
  \bibinfo{author}{M.~Tan}, \bibinfo{author}{A.~Casanova},
  \bibinfo{author}{M.~Labinaz}, \bibinfo{author}{K.~Sridhar},
  \bibinfo{author}{D.~H. Fitchett}, \bibinfo{author}{A.~Langer},
  \bibinfo{author}{S.~G. Goodman},
\newblock \bibinfo{title}{{Risk scores for risk stratification in acute
  coronary syndromes: useful but simpler is not necessarily better}},
\newblock \bibinfo{journal}{European Heart Journal} \bibinfo{volume}{28}
  (\bibinfo{year}{2007}) \bibinfo{pages}{1072--1078}.
  \DOIprefix\doi{10.1093/eurheartj/ehm004}.
\bibitem[{Mehta et~al.(2009)Mehta, Granger, Boden, Steg, Bassand, Faxon, Afzal,
  Chrolavicius, Jolly, Widimsky, Avezum, Rupprecht, Zhu, Col, Natarajan,
  Horsman, Fox, and Yusuf}]{Mehta2009}
\bibinfo{author}{S.~R. Mehta}, \bibinfo{author}{C.~B. Granger},
  \bibinfo{author}{W.~E. Boden}, \bibinfo{author}{P.~G. Steg},
  \bibinfo{author}{J.-P. Bassand}, \bibinfo{author}{D.~P. Faxon},
  \bibinfo{author}{R.~Afzal}, \bibinfo{author}{S.~Chrolavicius},
  \bibinfo{author}{S.~S. Jolly}, \bibinfo{author}{P.~Widimsky},
  \bibinfo{author}{A.~Avezum}, \bibinfo{author}{H.-J. Rupprecht},
  \bibinfo{author}{J.~Zhu}, \bibinfo{author}{J.~Col}, \bibinfo{author}{M.~K.
  Natarajan}, \bibinfo{author}{C.~Horsman}, \bibinfo{author}{K.~A. Fox},
  \bibinfo{author}{S.~Yusuf},
\newblock \bibinfo{title}{{Early versus Delayed Invasive Intervention in Acute
  Coronary Syndromes}},
\newblock \bibinfo{journal}{New England Journal of Medicine}
  \bibinfo{volume}{360} (\bibinfo{year}{2009}) \bibinfo{pages}{2165--2175}.
  \DOIprefix\doi{10.1056/NEJMoa0807986}.
\bibitem[{van Stralen et~al.(2009)van Stralen, Stel, Reitsma, Dekker, Zoccali,
  and Jager}]{VanStralen2009}
\bibinfo{author}{K.~J. van Stralen}, \bibinfo{author}{V.~S. Stel},
  \bibinfo{author}{J.~B. Reitsma}, \bibinfo{author}{F.~W. Dekker},
  \bibinfo{author}{C.~Zoccali}, \bibinfo{author}{K.~J. Jager},
\newblock \bibinfo{title}{{Diagnostic methods I: sensitivity, specificity, and
  other measures of accuracy}},
\newblock \bibinfo{journal}{Kidney International} \bibinfo{volume}{75}
  (\bibinfo{year}{2009}) \bibinfo{pages}{1257--1263}.
  \DOIprefix\doi{10.1038/ki.2009.92}.
\bibitem[{Mandrekar(2010)}]{Mandrekar2010}
\bibinfo{author}{J.~N. Mandrekar},
\newblock \bibinfo{title}{{Simple Statistical Measures for Diagnostic Accuracy
  Assessment}},
\newblock \bibinfo{journal}{Journal of Thoracic Oncology} \bibinfo{volume}{5}
  (\bibinfo{year}{2010}) \bibinfo{pages}{763--764}.
  \DOIprefix\doi{10.1097/JTO.0b013e3181dab122}.
\bibitem[{Schetinin et~al.(2018)Schetinin, Jakaite, and
  Krzanowski}]{Schetinin2018}
\bibinfo{author}{V.~Schetinin}, \bibinfo{author}{L.~Jakaite},
  \bibinfo{author}{W.~Krzanowski},
\newblock \bibinfo{title}{{Bayesian averaging over Decision Tree models for
  trauma severity scoring}},
\newblock \bibinfo{journal}{Artificial Intelligence in Medicine}
  \bibinfo{volume}{84} (\bibinfo{year}{2018}) \bibinfo{pages}{139--145}.
  \DOIprefix\doi{10.1016/j.artmed.2017.12.003}.

\end{thebibliography}
\end{document}